\documentclass{article} 
\usepackage{iclr2025_conference,times}

\usepackage{amsmath,amsfonts,bm}

\def\eqref#1{equation~\ref{#1}}

\def\1{\bm{1}}

\DeclareMathAlphabet{\mathsfit}{\encodingdefault}{\sfdefault}{m}{sl}
\SetMathAlphabet{\mathsfit}{bold}{\encodingdefault}{\sfdefault}{bx}{n}

\DeclareMathOperator*{\argmin}{arg\,min}

\usepackage{hyperref}
\usepackage{url}
\usepackage[utf8]{inputenc} 
\usepackage[T1]{fontenc}    
\usepackage{graphicx}
\usepackage{booktabs}
\usepackage{enumitem}
\usepackage{multirow}
\usepackage{algorithm, setspace} 
\usepackage{algorithmicx, algpseudocode}
\usepackage{booktabs}       
\usepackage{amsfonts}       
\usepackage{nicefrac}       
\usepackage{microtype}      
\usepackage{xcolor}         
\usepackage{orcidlink}      
\usepackage[accsupp]{axessibility}  
\usepackage{comment}
\usepackage{xcolor}
\usepackage[utf8]{inputenc}
\usepackage{graphicx}
\usepackage{xcolor}         
\usepackage{amsmath}
\usepackage{amssymb}
\usepackage{amsthm}
\usepackage{bm}
\usepackage{bbm}
\usepackage{nicefrac}
\usepackage{mathtools}
\usepackage{tablefootnote}
\usepackage{setspace}
\usepackage{subcaption}
\usepackage{wrapfig} 
\usepackage{blindtext}
\usepackage{pifont}
\usepackage[capitalize]{cleveref}
\usepackage{booktabs}

\title{Growing Efficient Accurate and Robust \\Neural Networks on the Edge}

\author{Vignesh Sundaresha \& Naresh Shanbhag \\
University of Illinois at Urbana-Champaign\\
Urbana, USA \\
\texttt{\{vs49, shanbhag\}@illinois.edu} \\
\vspace{-30pt}
}

\algrenewcommand{\algorithmiccomment}[1]{\textcolor{blue}{/* #1 */}}
\newcommand{\algcomment}[1]{\textcolor{blue}{// #1}}

\iclrfinalcopy 
\begin{document}

\maketitle

\begin{abstract}
    The ubiquitous deployment of deep learning systems on resource-constrained Edge devices is hindered by their high computational complexity coupled with their fragility to out-of-distribution (OOD) data, especially to naturally occurring common corruptions. Current solutions rely on the Cloud to train and compress models before deploying to the Edge. This incurs high energy and latency costs in transmitting locally acquired field data to the Cloud while also raising privacy concerns. We propose GEARnn (Growing Efficient, Accurate, and Robust neural networks) to grow and train robust networks in-situ, i.e., completely on the Edge device. Starting with a low-complexity initial backbone network, GEARnn employs One-Shot Growth (OSG) to grow a network satisfying the memory constraints of the Edge device using clean data, and robustifies the network using Efficient Robust Augmentation (ERA) to obtain the final network. We demonstrate results on a NVIDIA Jetson Xavier NX, and analyze the trade-offs between accuracy, robustness, model size, energy consumption, and training time. Our results demonstrate the construction of efficient, accurate, and robust networks entirely on an Edge device.
\end{abstract}

\vspace{-10pt}
\section{Introduction}
\label{sec: intro}

\begin{wrapfigure}{R}{0.5\textwidth}
        \centering
        \vspace{-15pt}
        \includegraphics[width=\linewidth]{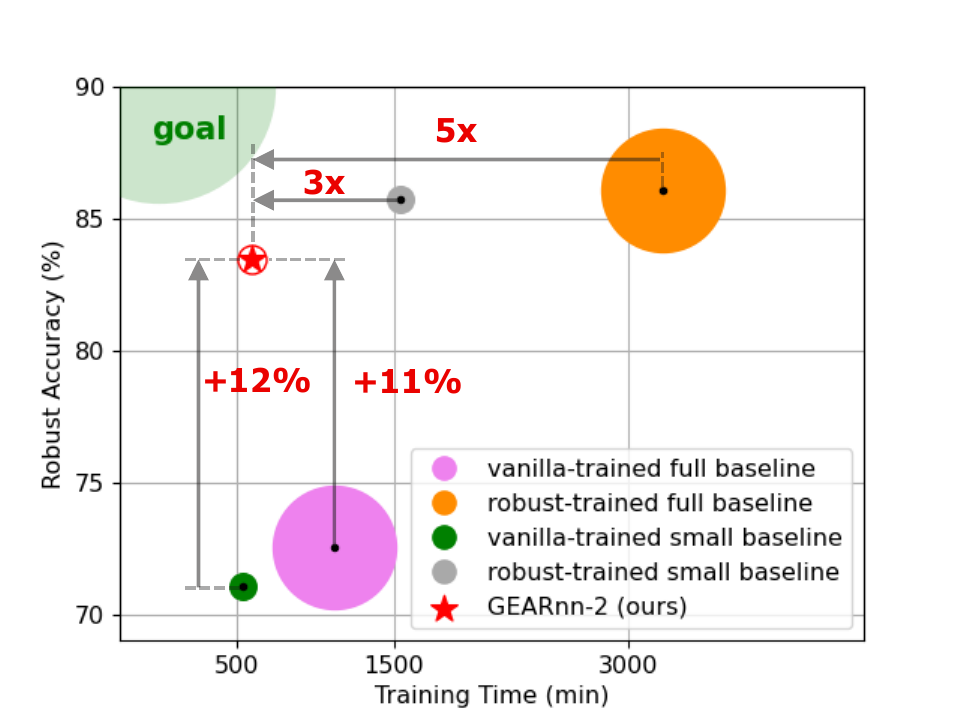}
        \caption{\label{fig: intro plot}Improvements in robust accuracy, training time, and model size (area of circles) of our proposed GEARnn
        method measured on NVIDIA Jetson Xavier NX Edge device~\cite{jetson}. Robust accuracy is evaluated on CIFAR-10-C for GEARnn, full network baselines (VGG-19), and small network baselines (5\% VGG-19 networks with same topology as GEARnn-2). For robust training, we employ AugMix~\cite{hendrycks2019augmix}. GEARnn demonstrates significant reduction in training complexity over robust baselines at similar robust and clean accuracies (shown in~\cref{subsec: Jetson results}).
        \vspace{-20pt}
        }
  \end{wrapfigure}

The ubiquitous practical deployment of deep neural networks is mainly hindered by their lack of robustness and high computational cost. Prior art has shown that these deep networks are extremely fragile to adversarial perturbations~\cite{szegedy2013intriguing}\cite{goodfellow2014explaining} and out-of-distribution (OOD) data~\cite{hendrycks2019benchmarking}\cite{mintun2021interaction}. Natural corruptions~\cite{hendrycks2019benchmarking} (a specific type of OOD data) are more commonly encountered at the Edge where real-time data is being continually acquired, e.g., video sequences acquired by on-board cameras in autonomous agents (self-driving cars, field robots, drones), which tend to be distorted by weather and blur. The state-of-the-art defense against these corruptions employs robust data augmentation~\cite{hendrycks2019augmix, hendrycks2021many, modas2022prime} which incurs a huge computational cost when implemented on an Edge device. \cref{fig: intro plot} indicates that it takes more than 2 days to robustly train a VGG-19 network~\cite{simonyan2014very} on a simple CIFAR-10 dataset when implemented on NVIDIA Jetson Xavier NX Edge device~\cite{jetson}. Even for a small 5\% VGG-19 network it takes more than a day, thus highlighting the non-trivial nature of the problem. This is a huge concern because Edge devices are typically battery-powered and such large training costs reduce their operational life-time.


Traditional solutions for reducing network complexity such as pruning~\cite{han2015learning, li2016pruning, diffenderfer2021winning}, quantization~\cite{rastegari2016xnor, hubara2016binarized} and neural architecture search (NAS)~\cite{liu2018progressive, zoph2018learning} mainly target Edge inference, and are not suited for Edge training since they start with hard-to-fit over-parameterized networks that require the large computational resources of the Cloud. However, transmitting local data to the Cloud incurs energy and latency costs while raising privacy concerns, thus requiring training to happen completely on the Edge.
Given the above challenges, the primary objective of our work is: \textit{To design and train compact robust networks completely in-situ on Edge devices.} Our proposed solution GEARnn (Growing Efficient, Accurate, and Robust neural networks) is based on the family of growth algorithms~\cite{chen2015net2net,wu2020firefly,evci2022gradmax,yuan2020growing} that gradually increase the size of an initial backbone network to reach the robust accuracy of a full network but at a fraction of its size, training complexity, and energy consumption. 

Prior work on network growth~\cite{wu2020firefly, wu2019splitting, yuan2023accelerated} do not consider robustness to common corruptions since they use clean data during training, while works   
that consider robustness train fixed-sized networks using augmented data~\cite{hendrycks2019augmix, modas2022prime} without considering the efficiency of robust training. 
Hence, in order to grow robust networks on the Edge and achieve good robustness vs. training efficiency trade-off, we ask the following questions: \textbf{Q1)} \textit{should networks be grown using augmented data only (1-Phase), or should they be grown using clean data first and then trained with augmented data (2-Phase)?} \textbf{Q2)} \textit{for growth, how many steps should be employed?}
We answer these questions by proposing our method GEARnn to efficiently grow robust networks. \cref{fig: intro plot} shows that GEARnn achieves significant improvements in robust accuracy over vanilla trained baselines while requiring much smaller training energy consumption compared to robustly trained baselines.

\begin{wrapfigure}{R}{0.7\textwidth}
        \centering
        \includegraphics[width=\linewidth]{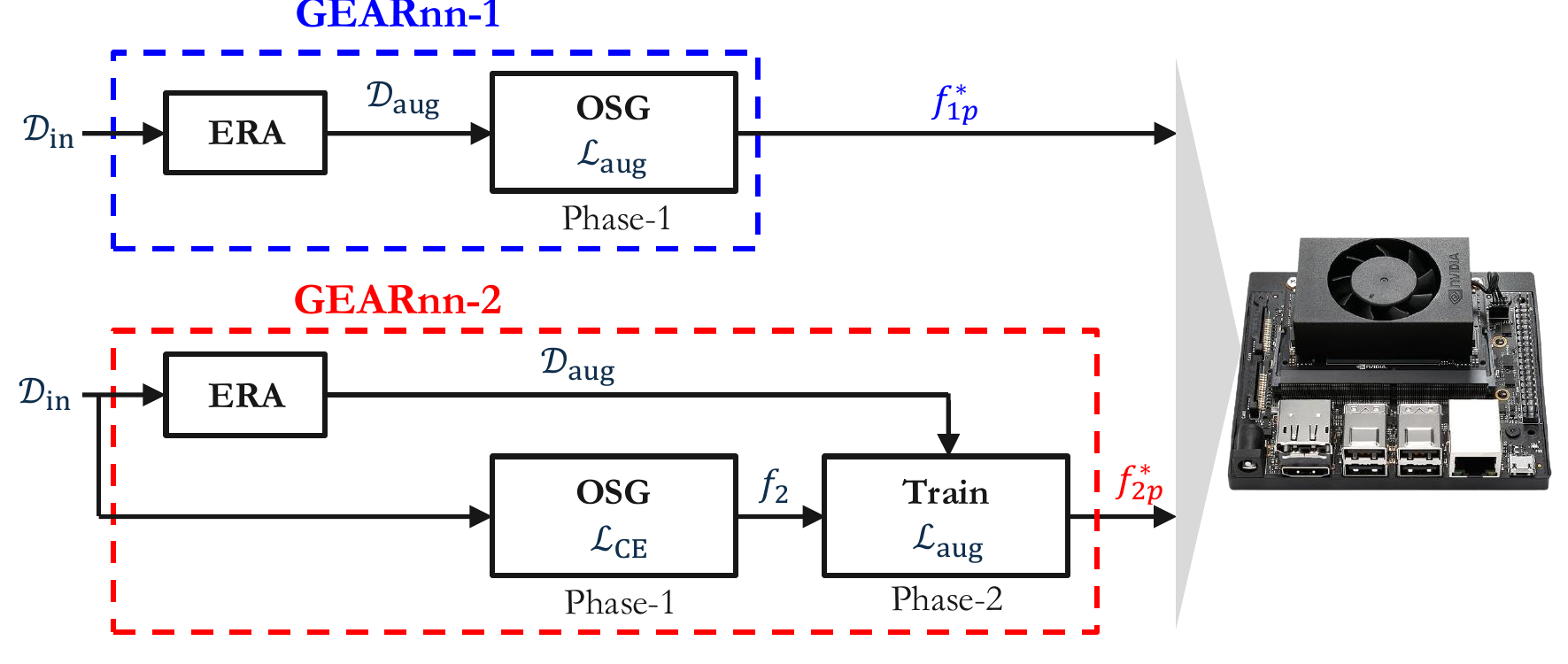}
        \caption{\label{fig: vision} Proposed approach: GEARnn-1 performs One-Shot Growth (OSG) on augmented data ($\mathcal{D}_{\text{aug}}$) generated by Efficient Robust Augmentation (ERA) (using clean data ($\mathcal{D}_{\text{in}}$)) in a single phase (1-Phase). GEARnn-2 performs OSG using $\mathcal{D}_{\text{in}}$ first followed by parametric training on $\mathcal{D}_{\text{aug}}$ in two consecutive phases
        (2-phase). Here $\mathcal{L}_{\text{CE}}$ and $\mathcal{L}_{\text{aug}}$ denote the cross-entropy loss and augmented loss, respectively.
        \vspace{-7pt}
        }
  \end{wrapfigure}

\textbf{Contributions}: We make the following contributions (\cref{fig: vision}):
\begin{enumerate}
    \item To the best of our knowledge, our work is the first to \textit{grow} networks robust to common corruptions.
  Additionally, we also consider the training efficiency of our methods.
\end{enumerate}
\vspace{-10pt}

\begin{enumerate}[resume, start=2]
    \item We answer \textbf{Q1} as: 2-Phase (growth with clean data followed by robust training using augmented data) provides improved robustness over a 1-Phase (growth using augmented data) at iso-model size. This result indicates the importance of proper initialization for efficient robust training (Sections~\ref{subsec: Quadro GEARnn},~\ref{subsec: Jetson results} \&~\ref{subsec: growth rationale}). 
   \item We answer \textbf{Q2} as: One-Shot Growth (OSG) achieves the best training efficiency, clean and robust accuracies at iso-model size compared to $m$-Shot ($m>1$) Growth (Section~\ref{subsec: growth rationale}).
    \item We propose two GEARnn (Growing Efficient Accurate and Robust neural networks) algorithms (see~\cref{fig: vision} and \cref{sec: Algorithm}) by combining 1-Phase/2-Phase  with OSG and Efficient Robust Augmentation (ERA).
    \item We explain the network topologies generated during OSG, and also provide rationale for the efficacy of 2-Phase approach and initialization using clean data (\cref{sec: Discussion}).
    \item We show that GEARnn generated networks shine on all four metrics simultaneously -- clean accuracy, robust accuracy, training efficiency and inference efficiency -- by implementing them on a real-life Edge device, the NVIDIA Jetson Xavier NX 
    (\cref{subsec: Jetson results}).
    
\end{enumerate}
\vspace{-10pt}
\section{Background and Related Work}
\label{sec:relatedworks}
\vspace{-10pt}

\textbf{Robust Data Augmentation:}
This is the most commonly used method for addressing corruptions due to its ease of integration into the training flow and ability to replicate low-level structural distortions. AugMix~\cite{hendrycks2019augmix}, PRIME~\cite{modas2022prime} and FourierMix~\cite{sun2021certified} combine chains of stochastic image transforms and enforce consistency using a suitable loss function to generate an augmented sample from a clean image. DeepAugment~\cite{hendrycks2021many} randomly distorts the parameters of an image-to-image network to generate augmented images. CARDs~\cite{diffenderfer2021winning} combines data augmentation~\cite{hendrycks2019augmix} and pruning~\cite{frankle2018lottery} to find compact robust networks embedded in large over-parameterized networks. Adversarial augmentations~\cite{zhao2020maximum, rusak2020increasing, calian2021defending} have also been proposed to handle common corruptions. 
Unlike our proposed GEARnn algorithm, all these techniques significantly increase the complexity over vanilla training and are thus inappropriate for Edge deployment.

\textbf{Growth Techniques:}
A typical growth algorithm starts with a small initial backbone model whose size is gradually increased until the desired performance or network topology is reached. Neural network growth has been previously used in optimization~\cite{fukumizu2000local}, continual learning~\cite{rusu2016progressive, hung2019compacting} and in speeding up the training of large networks~\cite{chen2015net2net}. Recent works~\cite{evci2022gradmax, yuan2023accelerated} look at improving the training dynamics and efficiency for growth by using better neuron initializations. 
Others find efficient networks by growing the width~\cite{wu2019splitting}, depth~\cite{wen2020autogrow} or both~\cite{wu2020firefly, yuan2020growing}. However, none of these methods address the issue of robustness to common corruptions or demonstrate the utility for training on a resource-constrained Edge setting, which is our focus. Though our work GEARnn builds upon Firefly~\cite{wu2020firefly}, it is flexible and can incorporate other growth methods mentioned above.
\vspace{-10pt}
\section{Notation and Problem Setup}
\label{sec: Notation}
\textbf{Notation: }Let $f:\mathbb{R}^d\rightarrow[C]$ be a hard classifier which classifies input $\mathbf{x}\in \mathbb{R}^d$ into one of $C$ classes. We choose $f$ to be a convolutional neural network (CNN) with $L$ layers (depth), $\{w_l\}_{l=1}^L$ output channels (widths), and $(K, K)$ sized kernels. The network $f$ is trained on $n$ samples $(\mathbf{x}, y)\sim\mathcal{D}_\text{in}$, where $(\mathbf{x}, y)\in \mathbb{R}^d\times[C]$ and $\mathcal{D}_{\text{in}}$ denotes the ``in-distribution'' or ``clean'' data. $\mathcal{L}_{\text{CE}}$ represents the cross-entropy loss and $\mathcal{L}_{\text{aug}} = \mathcal{L}_{\text{CE}} + \lambda\mathcal{L}_{\text{JSD}}$ represents the augmentation loss where $\mathcal{L}_{\text{JSD}}$ is the Jensen-Shannon divergence loss described in~\cite{hendrycks2019augmix}. 

During inference, $f$ can be exposed to samples from both $\mathcal{D}_{\text{in}}$ and $\mathcal{D}_{\text{out}}$ (``out-of-distribution'' or ``corrupted'' data). In case of common corruptions, $(\mathbf{x_{\text{out}}},y)\sim\mathcal{D}_{\text{out}}$ is obtained by $\mathbf{x_{\text{out}}} = \kappa(\mathbf{x_{\text{in}}},s)$, where $(\mathbf{x_{\text{in}}},y)\sim\mathcal{D}_{\text{in}}$, $\kappa$ is a corruption filter and $s$ is the severity level of the corruption. We denote $p_e=\text{Pr}(\hat{y}\neq y)$ as the classification error at inference where $\hat{y} = f(\mathbf{x}_\text{test})$. When $(\mathbf{x}_\text{test},y)\sim \mathcal{D}_{\text{in}}$, we define $(1-p_e)$ as  clean accuracy $\mathcal{A}_{\text{cln}}$, and when $(\mathbf{x}_\text{test},y)~\sim~\mathcal{D}_{\text{out}}$, we define $(1-p_e)$ as  robust accuracy $\mathcal{A}_{\text{rob}}$. The value of $p_e$ is determined empirically in this work.

\noindent\textbf{Problem: } Our primary objective is to maximize the empirical clean and robust accuracies ($\mathcal{A}_{\text{cln}}$ and $\mathcal{A}_{\text{rob}}$) while ensuring the network complexity ($\sum_{l=1}^Lw_l$) is small. Along with these two criteria, we also prioritize reduction in training time ($t_{\text{tr}}$) and training energy consumption ($E$) on hardware.
\vspace{-10pt}
\section{Growing Efficient Accurate and Robust Neural Networks (GEARnn)}
\label{sec: gearnn}
As shown in~\cref{fig: vision}, two flavors of GEARnn algorithms are proposed -- GEARnn-1 and GEARnn-2. While GEARnn-1 leverages the 1-Phase (joint growth and robust training) training, GEARnn-2 employs the 2-Phase (sequential growth and robust training) approach. Both flavors incorporate One-Shot Growth (OSG) and Efficient Robust Augmentation (ERA) in different ways. In this section, we first describe OSG and ERA, and then formally present the GEARnn algorithms.
\vspace{-5pt}
\subsection{One-Shot Growth (OSG)}
\label{subsec: OSG}
 One-Shot Growth (OSG) employs labeled data to perform a single growth step sandwiched between two training stages. The initial backbone $f_0$ is first trained for $\mathcal{E}_1$ epochs. The resulting network $f_1$ is grown over $\mathcal{E}_g$ epochs to obtain the grown network $f_g$, i.e., $f_g = \mathcal{G}(f_1|\gamma, \mathcal{D}, \mathcal{L}, \mathcal{E}_g)$, where $\mathcal{G}$ is the growth technique which is nominally Firefly~\cite{wu2020firefly} in our work. The final network $f_2$ is obtained by training $f_g$ over $\mathcal{E}_2$ epochs. Either clean ($\mathcal{D}_\text{in}$) or augmented ($\mathcal{D}_\text{aug}$) data can be used in OSG. For instance, OSG in GEARnn-1 and GEARnn-2 employs augmented data ($\mathcal{D} \sim \mathcal{D}_{\text{aug}}$) and clean data ($\mathcal{D} \sim \mathcal{D}_{\text{in}}$), respectively.
 
 The growth technique $\mathcal{G}$ is described below:
\begin{align}
\label{eqn: OSG}
\begin{split}
    \displaystyle f_g = \argmin_{f} &\hspace{1em}\mathcal{L}(f, \mathcal{D}|f_1) \\
    \text{s.t.} \hspace{1em} \hspace{1em}&f\in\partial(f_1,\epsilon)\\
    \mathcal{C}(f&) \leq (1+\gamma) \;\mathcal{C}(f_1)\
\end{split}
\end{align} 
where $\partial(f_1,\epsilon)$ represents the growth neighbourhood for topology search, $\mathcal{C}(f)=\sum_{l=1}^Lw_l$ represents the complexity estimate of network $f$ and $\gamma$ denotes the growth ratio. The neighbourhood $\partial(f_1,\epsilon)$ is expanded in two ways - splitting and growing new neurons - as described in \cite{wu2020firefly, wu2019splitting}. We perform growth only in the width dimension and keep the number of layers $L$ and the kernel size $(K, K)$ constant for reasons described in~\cite{wu2020firefly} and \cite{simonyan2014very}. 

Existing growth methods~\cite{wu2019splitting, wu2020firefly, evci2022gradmax} use several growth steps (typically 10 steps) and large number of training epochs (typically 1600 total epochs) which makes them inefficient for training. 
This directs us to pick OSG over multi-step growth (validated in~\cref{subsec: growth rationale}) and reduce the training epochs significantly ($20\times$) compared to prior growth algorithms. 
The drop in accuracy observed due to these modifications is compensated for using a 2-Phase approach (\cref{subsec: Quadro GEARnn}).
\begin{figure}
    \centering
    \includegraphics[width=\columnwidth]{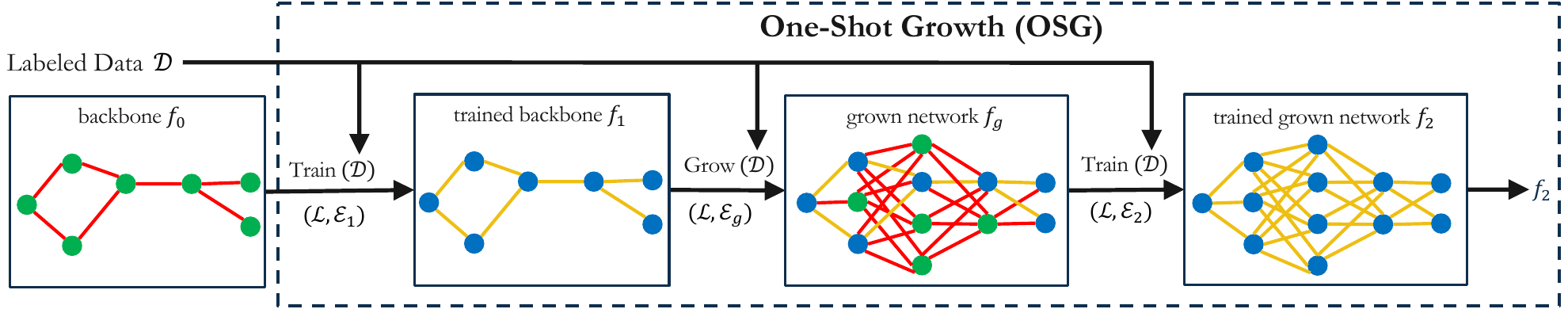}
    \caption{OSG takes in labeled data ($\mathcal{D}$) and backbone network $f_0$, and performs a training step, a growth step, and a training step in sequence to generate  network $f_2$. The 2-tuple $(\mathcal{L}, \mathcal{E})=$ (loss function, number of epochs) employed in each step.
    \vspace{-15pt}}
    \label{fig: OSG}
\end{figure}
\vspace{-10pt}
\subsection{Efficient Robust Augmentation (ERA)}
\label{subsec: efficient AugMix}
\begin{wrapfigure}{R}{0.55\textwidth}
        \vspace{-20pt}
        \centering
        \includegraphics[width=\linewidth]{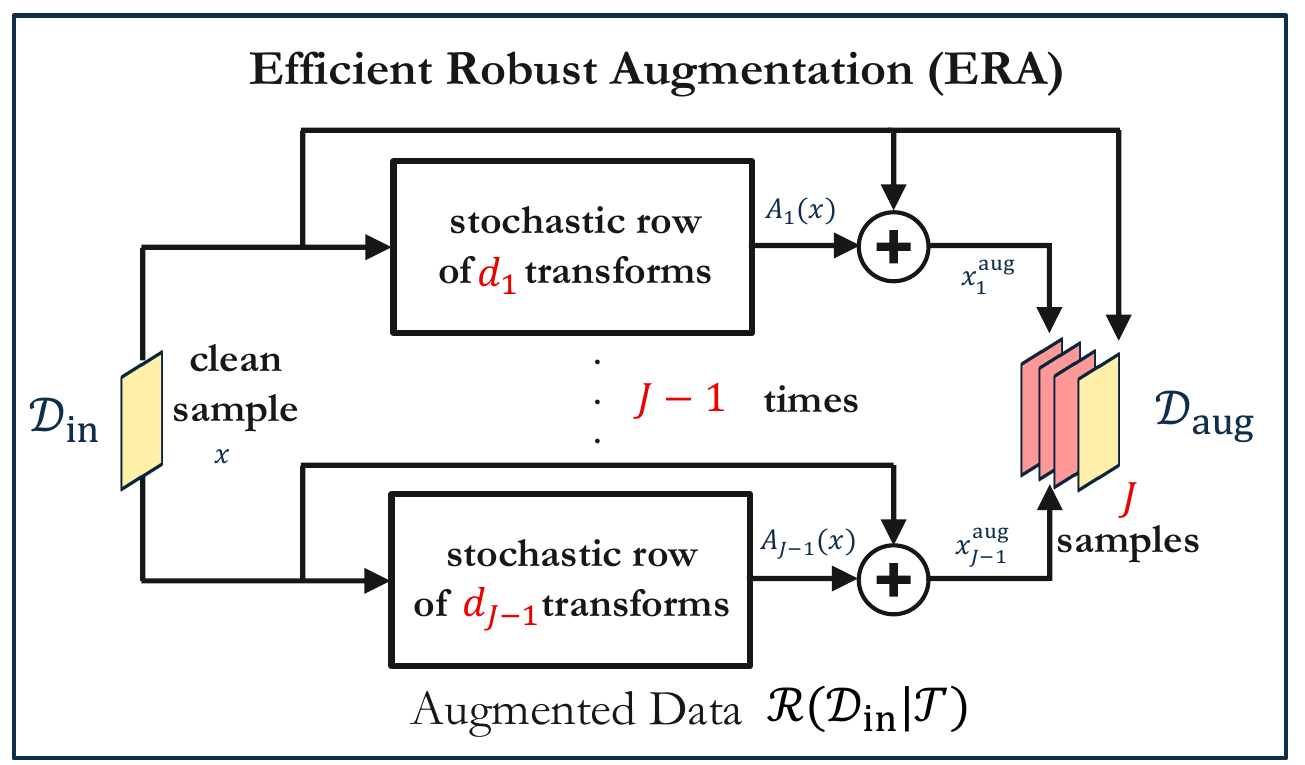}
        \caption{\label{fig: ERA} ERA takes in clean data ($\mathcal{D}_{\text{in}}$) as input and applies a set of stochastic transforms to generate augmented data ($\mathcal{D}_{\text{aug}}$) in an efficient manner.
        }
\end{wrapfigure}
Efficient Robust Augmentation (ERA) employs clean data ($\mathcal{D}_{\text{in}}$) to generate augmented data ($\mathcal{D}_{\text{aug}}$) in an efficient manner. The clean sample $\mathbf{x}$ (where ($\mathbf{x}, y)\sim \mathcal{D}_{\text{in}}$) is passed through a set of transforms $a_1, a_2, ... ,a_{d_{j}}$ to obtain the transformed sample $A_j(x)$, which is then combined linearly with the clean sample to give the augmented sample $\mathbf{x}_{j}^{\text{aug}}$. We concatenate $(J-1)$ such augmented samples $\{\mathbf{x}^{\text{aug}}_j\}_{j=1}^{J-1}$ along with the clean sample to obtain our Efficient Robust Augmentation $\mathcal{R} ((\mathbf{x},y)|\mathcal{T})$.\\
\begin{align}\label{eqn: ERA}
    \begin{split}
        A_j(\mathbf{x}) &= a_1\circ a_2\circ ... \circ a_{d_{j}}(\mathbf{x})\\
        \mathbf{x}_j^{\text{aug}} &= p\mathbf{x} + (1-p)A_j(\mathbf{x})\\
        \mathcal{R}((\mathbf{x},y)|\mathcal{T}) = (\{\mathbf{x}_1^{\text{aug}}, &..., \mathbf{x}_{J-1}^{\text{aug}}, \mathbf{x}\},y) \implies  \mathcal{D_{\text{aug}}}:= \mathcal{R} (\mathcal{D}_{\text{in}}|\mathcal{T})\\
        \text{where } a_i \sim \text{Unif}(\mathcal{T}),\; p\sim \beta(1,1)&    ,\;d_j\sim\text{Unif}(\{1,...,D\}),\; j\in\{1,...,J-1\}
    \end{split}
\end{align}
 where $\mathcal{T}$ denotes the set of transforms, $\beta()$ and Unif() represent the beta and uniform distributions, respectively. SOTA robust data augmentation~\cite{hendrycks2019augmix, modas2022prime, sun2021certified} methods for common corruptions also employ stochastic chains of transforms with width $W$, depth $D$, and enforce consistency across $J-1$ augmented and clean samples using the $\mathcal{L}_{\text{aug}}$ loss function (described in~\cref{sec: Notation}). The SOTA augmentation framework increases the training time and energy by $3\times$$\text{ to }4\times$ compared to vanilla training. We choose $(W,D,J) = (1,3,4)$ based on our diagnosis (shown in~\cref{app: ERA diagnosis}) to improve the efficiency without compromising on robustness compared to SOTA approaches~\cite{hendrycks2019augmix, modas2022prime}.
 In GEARnn-2, grown network $f_2$ (see Fig.~\ref{fig: OSG}) obtained using clean data OSG is trained for $\mathcal{E}_r$ epochs using $\mathcal{D}_{\text{aug}}$ generated by ERA.
\vspace{-5pt}
\subsection{GEARnn Algorithms}
\label{sec: Algorithm}
\begin{wrapfigure}{L}{0.56\textwidth}
    \vspace{-8pt}
    \scalebox{0.7}{
    \begin{minipage}{0.78\textwidth}
        \begin{algorithm}[H]
        \setstretch{1.3}
           \caption{GEARnn-1}
           \label{alg:gearnn1}
        \begin{algorithmic}[1]
           \State {\bfseries Input:} clean training data $\mathcal{D}_{\text{in}}$, initial backbone network $f_0$, growth ratio $\gamma$, set of augmentation transforms $\mathcal{T}$, training epochs $\{\mathcal{E}_1, \mathcal{E}_g, \mathcal{E}_2\}$
           \State {\bfseries Output:} compact and robust model $f_{1p}^*$
           \State \Comment{Phase-1: OSG}
               \For{$e = 1,...,\mathcal{E}_1$}
                    \State $\mathcal{D_{\text{aug}}}:= \mathcal{R} (\mathcal{D}_{\text{in}}|\mathcal{T})$\quad\algcomment{ERA}
                    \State $f_1 \leftarrow \displaystyle\argmin_f \hspace{0.5em}\mathcal{L}_{\text{aug}}(f, \mathcal{D}_{\text{aug}} | f_0)$ \quad\algcomment{backbone robust training}
                \EndFor
           \State $f_g \leftarrow \mathcal{G}(f_1|\gamma, \mathcal{D}_{\text{aug}}, \mathcal{L}_{\text{aug}}, \mathcal{E}_g)$ \quad\algcomment{augmented growth}
               \For{$e = 1,...,\mathcal{E}_2$}
                    \State $\mathcal{D_{\text{aug}}}:= \mathcal{R} (\mathcal{D}_{\text{in}}|\mathcal{T})$\quad\algcomment{ERA}
                    \State $f_2 \leftarrow \displaystyle\argmin_f \hspace{0.5em}\mathcal{L}_{\text{aug}}(f, \mathcal{D}_{\text{aug}} | f_g)$ \quad\algcomment{grown-network robust training}
                \EndFor
                \State $f_{1p}^* \leftarrow f_2$
                \State {\textbf{return}} $f_{1p}^*$
        \end{algorithmic}
        \end{algorithm} 
        \begin{algorithm}[H]
        \setstretch{1.3}
           \caption{GEARnn-2}
           \label{alg:gearnn}
        \begin{algorithmic}[1]
           \State {\bfseries Input:} clean training data $\mathcal{D}_{\text{in}}$, initial backbone network $f_0$, growth ratio $\gamma$, set of augmentation transforms $\mathcal{T}$, training epochs $\{\mathcal{E}_1, \mathcal{E}_g, \mathcal{E}_2, \mathcal{E}_r\}$
           \State {\bfseries Output:} compact and robust model $f_{2p}^*$
           \State \Comment{Phase-1: OSG}
               \For{$e = 1,...,\mathcal{E}_1$}
                    \State $f_1 \leftarrow \displaystyle\argmin_f \hspace{0.5em}\mathcal{L}_{\text{CE}}(f, \mathcal{D}_{\text{in}} | f_0)$ \quad\algcomment{backbone clean training}
                \EndFor
           \State $f_g \leftarrow \mathcal{G}(f_1|\gamma, \mathcal{D}_{\text{in}}, \mathcal{L}_{\text{CE}}, \mathcal{E}_g)$ \quad\algcomment{clean growth}
               \For{$e = 1,...,\mathcal{E}_2$}
                    \State $f_2 \leftarrow \displaystyle\argmin_f \hspace{0.5em}\mathcal{L}_{\text{CE}}(f, \mathcal{D}_{\text{in}} | f_g)$ \quad\algcomment{grown-network clean training}
                \EndFor
           \State \Comment{Phase-2: Train}
            \For{$e = 1,...,\mathcal{E}_r$}
                    \State $\mathcal{D_{\text{aug}}}:= \mathcal{R} (\mathcal{D}_{\text{in}}|\mathcal{T})$\quad\algcomment{ERA}
                    \State $f_{2p}^* \leftarrow \displaystyle\argmin_f \hspace{0.5em}\mathcal{L}_{\text{aug}}(f, \mathcal{D_{\text{aug}}} | f_2)$ \algcomment{grown-network robust training}
                \EndFor
                \State {\textbf{return}} $f_{2p}^*$
        \end{algorithmic}
        \end{algorithm} 
    \end{minipage}
    }
    \vspace{-30pt}
  \end{wrapfigure}

  Algorithms~\ref{alg:gearnn1} and~\ref{alg:gearnn} describe GEARnn-1 and GEARnn-2, respectively. 
  Algorithms~\ref{alg:gearnn1} and~\ref{alg:gearnn} output final compact and robust models $f_{1p}^*$ and $f_{2p}^*$, respectively. 
  For empirical results in~\cref{sec: results}, the growth technique $\mathcal{G}$ and the set of transforms $\mathcal{T}$ are chosen from Firefly~\cite{wu2020firefly} and AugMix~\cite{hendrycks2019augmix}, respectively, though other growth~\cite{yuan2023accelerated, wu2019splitting} and augmentation~\cite{modas2022prime, sun2021certified} methods can be substituted to obtain different GEARnn variants. 
\vspace{-5pt}
\section{Experimental Setup}
\label{sec: setup}
\vspace{-10pt}
\textbf{Datasets and Architectures:}
All results are shown on CIFAR-10, CIFAR-100~\cite{krizhevsky2009learning} and Tiny ImageNet~\cite{le2015tiny} ($\mathcal{D}_{\text{in}}$) datasets. CIFAR-10-C, CIFAR-100-C and Tiny ImageNet-C~\cite{hendrycks2019benchmarking} ($\mathcal{D}_{\text{out}}$) are used to benchmark corruption robustness. Convolutional neural network architectures MobileNet-V1\cite{howard2017mobilenets}, VGG-19\cite{simonyan2014very}, ResNet-18\cite{he2016deep} are employed to demonstrate the results. 

\noindent\textbf{Hardware:}
For the server-based experiments, we use a single NVIDIA Quadro RTX 6000 GPU with 24GB RAM, 16.3 TFLOPS peak performance and an Intel Xeon Silver 4214R CPU. This machine is referred to as ``Quadro''. For the Edge-based experiments, we use the NVIDIA Jetson Xavier NX \cite{jetson} which has a Volta GPU with 8GB RAM, 21 TOPS peak performance and a Carmel CPU. We refer to this device as ``Jetson''.

\noindent\textbf{Metrics:}
Clean accuracy $\mathcal{A}_{\text{cln}} (\%)$ measured on clean test data $\mathcal{D}_{\text{in}}$, and robust accuracy $\mathcal{A}_{\text{rob}}(\%)$ measured on corrupted test data $\mathcal{D}_{\text{out}}$, are used as accuracy metrics (both computed using RobustBench~\cite{croce2021robustbench}). The number of floating-point parameters (model size), wall-clock training time $t_{\text{tr}}$ (in minutes), per-sample wall-clock inference time $t_{\text{inf}}$ (in seconds) and energy consumption $E$ (in Joule) are used as the efficiency metrics. Size $(\%)$ represents the fraction of the full model size. In case of growth algorithms, training times include both the time taken for training and growth. The power is measured from the Quadro and Jetson using Nvidia-SMI~\cite{nvidia-smi} and Jetson Stats~\cite{statsjetson}, respectively, and the energy $E$ is computed by summing the mean power values polled.

\noindent\textbf{Baselines:} In the absence of prior work on robust growth, we propose our own baselines Small ($\mathcal{D}_{\text{in}}$) and Small ($\mathcal{D}_{\text{aug}}$), both of which use 160 training epochs to be consistent with~\cite{diffenderfer2021winning}.
     They are networks with the same size and topology as the final GEARnn-2 network ($f_{2p}^*$ in Fig.~\ref{fig: vision}) trained with random initialization on clean data and augmented data (AugMix~\cite{hendrycks2019augmix}, unless specified otherwise), respectively.

We pick Small ($\mathcal{D}_{\text{aug}}$) as the main baseline for a fair comparison with GEARnn as it depicts a typical private-Edge training scenario. We do not compare with compression techniques since they have been shown to have worse training efficiency compared to growth~\cite{yuan2020growing}, and require a robust-trained full baseline, and this is  clearly more expensive than training Small ($\mathcal{D}_{\text{aug}}$) (see ~\cref{fig: intro plot}).

\begin{wraptable}{R}{0.6\columnwidth}
        \centering
        \vspace{-10pt}
        \setlength{\tabcolsep}{3pt}
        \caption{GEARnn hyperparameters for different networks and datasets.}
        \label{tab: gearnn hyperparams}          
        \centering
            \resizebox{\linewidth}{!}{
             \begin{tabular}{c | c c c | c | c c c | c c c c}
                    \toprule
                    \multirow{2}{*}{Dataset}  & \multicolumn{3}{c|}{Growth Ratio ($\gamma$)} & Small($\mathcal{D}$) & \multicolumn{3}{|c|}{GEARnn-1} & \multicolumn{4}{|c}{GEARnn-2}\\
                    & Mob. & VGG & Res. & $\mathcal{E}$ & $\mathcal{E}_1$ & $\mathcal{E}_g$ & $\mathcal{E}_2$ & $\mathcal{E}_1$ & $\mathcal{E}_g$ & $\mathcal{E}_2$ & $\mathcal{E}_r$\\
                    \midrule
                    CIFAR-10 & 1.8 & 0.9 & 0.6 & 160 & 40 & 1 & 40  & 40 & 1 & 40 & 40 \\
                    CIFAR-100 & 2.0 & 1.5 & 0.8 & 160 & 50 & 1 & 50 & 40 & 1 & 40 & 50\\
                    Tiny ImageNet & 2.0 & 1.5 & 0.8 & 160 & 50 & 1 & 50 & 40 & 1 & 40 & 50\\
                    \bottomrule
            \end{tabular}  
            }
        \vspace{-10pt}
\end{wraptable}

\vspace{-10pt}
\section{Main Results} 
\vspace{-5pt}
\label{sec: results}
In this section, we first compare the performance of GEARnn across different network architectures and datasets on Quadro. We then show results for CIFAR-10 and CIFAR-100 using VGG-19 and MobileNet on Jetson. Finally, we compare OSG with $m$-shot growth methods on Jetson.
\vspace{-5pt}
\subsection{Results across Network Architectures and Datasets}
\label{subsec: Quadro GEARnn}
\begin{wrapfigure}{R}{0.7\textwidth}
    \centering
    \begin{subfigure}{0.33\textwidth}
        \centering
        \includegraphics[width=\linewidth]{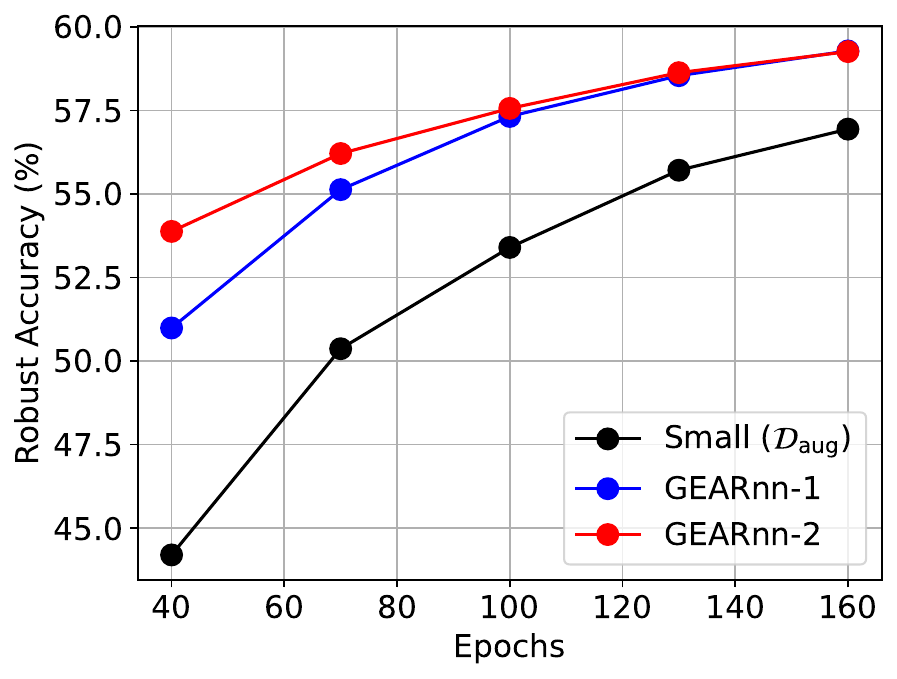}
        \subcaption{\label{epochs vs rob c100}}
        \label{fig:sub2}
    \end{subfigure}
    \begin{subfigure}{0.33\textwidth}
        \centering
            \includegraphics[width=\linewidth]{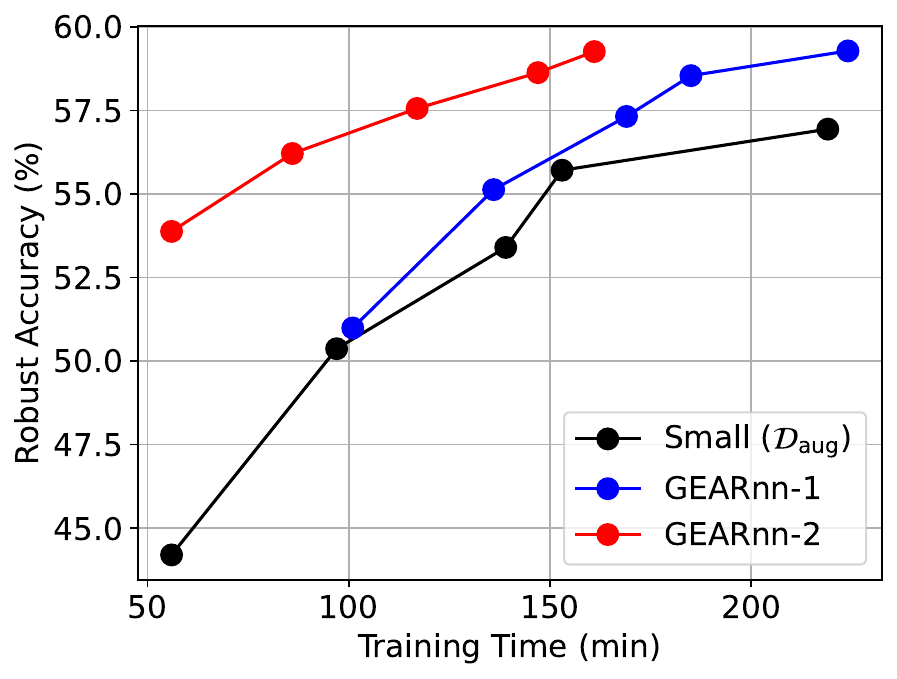}
        \subcaption{\label{ttr vs rob c100}}
        \label{fig: Arob_ttr}
    \end{subfigure}
    \caption{GEARnn-2 achieves higher robustness at the same: (a) number of robust training epochs at final model size, and (b) training time, for VGG-19/CIFAR-100 on Quadro.\vspace{-5pt}}
    \label{fig: robust epochs}        
  \end{wrapfigure}
\cref{tab: GEARnn Quadro} shows GEARnn is consistently better in terms of training time and training energy consumption over the best baseline Small ($\mathcal{D}_{\text{aug}}$) over multiple network architectures and datasets. Specifically, an average reduction in training time (energy consumption) of $\mathbf{3.5\times}$, $\mathbf{2.9\times}$ and $\mathbf{1.8\times}$ ($\mathbf{3.7\times}$, $\mathbf{2.0\times}$ and $\mathbf{2.0\times}$) is observed for CIFAR-10, CIFAR-100 and Tiny ImageNet, respectively. Furthermore, we find GEARnn-1 is inferior to GEARnn-2 on all the four metrics thereby answering \textbf{Q1} in~\cref{sec: intro} -- 2-Phase approach is better than 1-Phase approach for efficiently growing robust networks. 

A key reason underlying GEARnn-2's training efficiency is the reduction in the number of robust training epochs $\mathcal{E}_r$ made possible by the OSG initialization in Phase-1.~\cref{fig: robust epochs} shows that for the same training time, GEARnn-2 provides better robustness than Small ($\mathcal{D}_{\text{aug}}$) and GEARnn-1. Similar results were obtained for CIFAR-10 and other network architectures as shown in~\cref{app: robust epochs}.
\begin{table}[!t]
\centering
\caption{Comparison of accuracy, robustness, and efficiency between the baselines and GEARnn across various network architectures for CIFAR-10, CIFAR-100 and Tiny ImageNet on Quadro. See~\cref{fig: robust epochs} for robustness comparison between Small ($\mathcal{D}_{\text{aug}}$) and GEARnn-2 at similar training cost.
} 
\label{tab: GEARnn Quadro}
\setlength{\tabcolsep}{3pt}
\resizebox{\columnwidth}{!}{%
\begin{tabular}{c | c || c | c c | c c || c | c c | c c || c | c c | c c}

\toprule
Architecture & &\multicolumn{5}{c||}{CIFAR-10}&\multicolumn{5}{c||}{CIFAR-100}&\multicolumn{5}{c}{Tiny ImageNet}\\
\midrule
(full model & \multirow{2}{*}{Method} & Size & \multicolumn{2}{c|}{Accuracy}& \multicolumn{2}{c||}{Training} & Size & \multicolumn{2}{c|}{Accuracy}& \multicolumn{2}{c||}{Training} & Size & \multicolumn{2}{c|}{Accuracy}& \multicolumn{2}{c}{Training}\\

  size) & & (\%) & $\mathcal{A}_{\text{cln}}(\%)$ & $\mathcal{A}_{\text{rob}}(\%)$ & $t_\text{tr}$(min) & $E$(kJ) & (\%) & $\mathcal{A}_{\text{cln}}(\%)$ & $\mathcal{A}_{\text{rob}}(\%)$ & $t_\text{tr}$(min) & $E$(kJ) & (\%) & $\mathcal{A}_{\text{cln}}(\%)$ & $\mathcal{A}_{\text{rob}}(\%)$ & $t_\text{tr}$(min) & $E$(kJ)\\
\midrule
\midrule
            & Small ($\mathcal{D}_{\text{in}}$) & 8& 92.28 & 66.31\textcolor{red}{$\downarrow$} & 42 & 192 &
            8& 67.66 & 39.04\textcolor{red}{$\downarrow$} & 45 & 274 &
            8 & 55.13                & 18.48 \textcolor{red}{$\downarrow$}               & 262                & 2030\\ 
            
            \cmidrule{2-17}
            {MobileNetV1}  & Small ($\mathcal{D}_{\text{aug}}$) & 8 & \underline{\textbf{92.90}} & \underline{\textbf{83.21}} & 211 & 1130 &
            8 & \textbf{\underline{68.88}} & \textbf{\underline{54.95}} & 212 & 1330 &
            8 & \textbf{\underline{56.46}} & 28.17                & 765                & 7200 \\
            
            {(3M)}& GEARnn-1 &7 & 90.64 & 80.71 & \textbf{88} & \textbf{379} &
            8& 65.07 & 51.46 & \textbf{93} & \textbf{651} &
            8 & 54.57                & 27.46                & \textbf{506}       & \textbf{4410}\\
            
            & GEARnn-2 &8 & \textbf{91.35} & \textbf{81.96} & \underline{\textbf{56}} & \underline{\textbf{270}} &
            8& \textbf{67.95} & \textbf{53.28} & \textbf{\underline{72}} & \textbf{\underline{432}} &
            8 & \textbf{56.16}       & \textbf{\underline{28.56}} & \textbf{\underline{429}} & \textbf{\underline{3565}}\\
            
            \midrule 
            \midrule 
            
            & Small ($\mathcal{D}_{\text{in}}$) &5 & 92.69 & 70.57\textcolor{red}{$\downarrow$} & 31 & 241 & 
            9& 68.07 & 41.24\textcolor{red}{$\downarrow$} & 38 & 335 & 
            9 & 53.9                 & 17.78 \textcolor{red}{$\downarrow$}               & 218                & 2040\\
            
            \cmidrule{2-17}
            {VGG-19} & Small ($\mathcal{D}_{\text{aug}}$) &5 & \textbf{\underline{93.08}} & \textbf{\underline{85.73}} & 215 &  1140 &
            9 & \textbf{\underline{70.01}} & \textbf{\underline{56.94}} & 219 & 927 & 
            9 & \textbf{55.51}       & \textbf{\underline{30.01}} & 668                & 7120\\
            
            {(20M)} & GEARnn-1 &5 & 91.25 & 82.86 & \textbf{86} & \textbf{552}  &
            9& 65.73 & 52.68 & \textbf{111} & \textbf{779} & 
            9 & 54.38                & 28.56                & \textbf{428}       & \textbf{4220}\\
            
            & GEARnn-2 &5 & \textbf{92.18} & \textbf{83.77} & \textbf{\underline{53}} & \textbf{\underline{298}} &
            9& \textbf{68.44} & \textbf{54.31} & \textbf{\underline{65}} & \textbf{\underline{566}} & 
            9 & \textbf{\underline{56.19}} & \textbf{29.79}       & \textbf{\underline{357}} & \textbf{\underline{3219}}\\
            
            \midrule
            \midrule 
            
            & Small ($\mathcal{D}_{\text{in}}$) & 6& 93.34 & 68.85\textcolor{red}{$\downarrow$} & 61 & 546 & 
            7& 68.74 & 40.83\textcolor{red}{$\downarrow$} & 67 & 490 & 
            7 & \textbf{54.72}       & 18.11 \textcolor{red}{$\downarrow$}               & 381                & 3390\\

            \cmidrule{2-17}
            {ResNet-18} & Small ($\mathcal{D}_{\text{aug}}$) & 6 &\textbf{\underline{94.18}} & \textbf{\underline{86.50}} & 217 & 1730 &
            7 & \textbf{\underline{71.97}} & \textbf{\underline{57.30}} & 219 & 1250 & 
            7 & 54.50                 & \textbf{25.74}       & 1103               & 12400\\
            
            {(12M)}& GEARnn-1 & 6& 92.36 & 83.86 & \textbf{108} & \textbf{747}  &
            8& 69.15 & 55.62 & \textbf{142} & \textbf{1020} & 
            7 & 53.17                & 24.93                & \textbf{898}       & \textbf{9100}\\

            & GEARnn-2 & 6& \textbf{93.14} & \textbf{84.45} & \textbf{\underline{77}} & \textbf{\underline{567}} &
            7&\textbf{70.94} & \textbf{56.54} & \textbf{\underline{97}}& \textbf{\underline{905}} & 
            7 & \textbf{\underline{54.79}} & \textbf{\underline{26.64}} & \textbf{\underline{649}} & \textbf{\underline{7270}}\\
\bottomrule
\end{tabular}
}
\vspace{-10pt}
\end{table}
\vspace{-10pt}
\subsection{Results on the Edge}
\vspace{-5pt}
\label{subsec: Jetson results}
\begin{table}[!t]
\centering
\caption{Comparison of accuracy, robustness, inference and training efficiency between the baselines and GEARnn for CIFAR-10 and CIFAR-100 using MobileNet-V1 and VGG-19 on Jetson. Due to computational limitations, the results for Tiny ImageNet and ResNet-18 are excluded for Jetson.
} 
\label{tab: GEARnn Jetson}
\setlength{\tabcolsep}{3pt}
\resizebox{\columnwidth}{!}{%
\begin{tabular}{c || c || c c | c c | c c || c c | c c | c c}

\toprule
& & \multicolumn{6}{c||}{CIFAR-10}&\multicolumn{6}{c}{CIFAR-100}\\
\toprule
\multirow{2}{*}{Network} & \multirow{2}{*}{Method} & \multicolumn{2}{c|}{Accuracy}& \multicolumn{2}{c|}{Inference}&\multicolumn{2}{c||}{Training} & \multicolumn{2}{c|}{Accuracy}& \multicolumn{2}{c|}{Inference}&\multicolumn{2}{c}{Training}\\

  & & $\mathcal{A}_{\text{cln}}(\%)$ & $\mathcal{A}_{\text{rob}}(\%)$ & Size\% & $t_\text{inf}$(ms) & $t_\text{tr}$(min) & $E$(kJ) & $\mathcal{A}_{\text{cln}}(\%)$ & $\mathcal{A}_{\text{rob}}(\%)$ & Size\% & $t_\text{inf}$(ms) & $t_\text{tr}$(min) & $E$(kJ)\\
\midrule
\midrule
\multirow{4}{*}{MobileNet-V1} & Small ($\mathcal{D}_{\text{in}}$)  & 91.88 & 68.35\textcolor{red}{$\downarrow$} & 7 & 0.9 & 675 & 175 & 
68.59 & 39.47\textcolor{red}{$\downarrow$} & 8 & 0.9 & 744 & 166\\
\cmidrule{2-14} 
& Small ($\mathcal{D}_{\text{aug}}$) & \textbf{\underline{92.58}} & \textbf{\underline{83.84}} & 7 & 
\textbf{\underline{0.9}} & 1216 & 511 & 
\textbf{\underline{69.24}} & \textbf{\underline{54.84}} & 8 & \textbf{\underline{0.9}} & 1333 & 586\\
& GEARnn-1  & 90.20 & 79.65 & 7 & \textbf{\underline{0.9}}
& \textbf{{560}} & \textbf{238} 
& 65.48 & 50.46 & 8 & \textbf{1.0} & \textbf{704} & \textbf{{226}}\\
& GEARnn-2 & \textbf{91.43} & \textbf{81.64} & 7 &
\textbf{\underline{0.9}} & \textbf{\underline{553}} & \textbf{\underline{162}} & 
\textbf{67.42} & \textbf{52.39} & 8 & \textbf{\underline{0.9}} & \textbf{\underline{690}} & \textbf{\underline{216}}\\
\bottomrule
\midrule
\multirow{4}{*}{VGG-19} & Small ($\mathcal{D}_{\text{in}}$)  & 92.97 & 71.08\textcolor{red}{$\downarrow$} & 5 & 1.0 & 533&128 & 
67.92 & 40.49\textcolor{red}{$\downarrow$} & 9 & 1.4 & 714 & 187\\
\cmidrule{2-14}
&Small ($\mathcal{D}_{\text{aug}}$) & \textbf{\underline{93.36}} & \textbf{\underline{85.73}} & 5 & \textbf{\underline{1.0}} & 1543 & 522 & 
\textbf{\underline{70.07}} & \textbf{\underline{56.68}} & 9 & \textbf{\underline{1.4}} & 2016 & 678\\
&GEARnn-1  & 90.94 & 82.25 & 5 & \textbf{1.2} & \textbf{652} & \textbf{207} & 
62.89 & 49.63 & 9 & \textbf{1.5} & \textbf{936} & \textbf{\underline{281}}\\
&GEARnn-2 & \textbf{92.07} & \textbf{83.45} & 5& \textbf{\underline{1.0}} & \textbf{\underline{596}} & \textbf{\underline{155}} & 
\textbf{67.59} & \textbf{53.64} & 9 & \textbf{\underline{1.4}} & \textbf{\underline{884}} & \textbf{328}\\
\bottomrule
\end{tabular}
}
\vspace{-15pt}
\end{table}

We now study GEARnn when mapped onto the Edge device NVIDIA Jetson Xavier NX. The training hyperparameters for Jetson are described in~\cref{app: train setup}. Results on Jetson (\cref{tab: GEARnn Jetson}) show similar trends to those on Quadro (\cref{tab: GEARnn Quadro}). 

Specifically, ~\cref{tab: GEARnn Jetson} shows that GEARnn-2 achieves comparable clean and robust accuracies to the baseline Small ($\mathcal{D}_{\text{aug}}$) but at a fraction of its training cost -- a $\mathbf{2.3\times}$ ($\mathbf{2.8\times}$) reduction in training time (training energy) when averaged across both networks and datasets. Additionally, GEARnn-2 beats GEARnn-1 on almost all metrics, again confirming our answer to \textbf{Q1} in favour of 2-Phase. Interestingly, GEARnn-2 achieves a clean accuracy within $1\%$ of Small ($\mathcal{D}_{\text{in}}$) at a similar training cost. 
These results confirm that it is possible to grow efficient and robust networks on the Edge.
\vspace{-5pt}
\subsection{One-Shot vs. Multi-Shot Growth}
\vspace{-5pt}
Since GEARnn employs OSG (One-Shot Growth) for growing networks, it begs the question if we are missing anything if multiple growth steps ($m$-Shot Growth) were to be permitted, i.e., question \textbf{Q2} from~\cref{sec: intro}. 
To answer this question, we compare the clean and robust accuracies along with training time and energy for different growth steps between GEARnn-1 and GEARnn-2 in~\cref{tab: growth power}. 
All $m$-Shot Growth methods start with the same initial backbone $f_0$ (1.4\% of full model size) and perform growth to reach $f_2$ (5\% of full model size) using different growth ratios.  All methods use VGG-19 model and perform 80 epochs parametric training during the growth phase. The experiments are done on CIFAR-10 data and the hardware measurements are taken from Jetson.


\label{subsec: growth rationale}
\begin{wraptable}{R}{0.7\columnwidth}
    
        \vspace{-5pt}
        \setlength{\tabcolsep}{3pt}     
        \caption{Comparison of training complexities, clean and robust accuracies for different growth methods implemented using VGG-19 and CIFAR-10 on Jetson. 2-Phase approach and OSG provide the best solution for growing robust networks on the Edge.
        }
        \label{tab: growth power}          
        \centering
            \resizebox{\linewidth}{!}{
             \begin{tabular}{c | c c c c | c c c c }
                    \toprule
                    Growth & \multicolumn{4}{c|}{GEARnn-1}&\multicolumn{4}{c}{GEARnn-2} \\
                    Steps & 
                     $\mathcal{A}_\text{cln}(\%)$ & $\mathcal{A}_\text{rob}(\%)$ & $t_\text{tr}$ (min) & $E$ (kJ)
                    & $\mathcal{A}_\text{cln}(\%)$ & $\mathcal{A}_\text{rob}(\%)$ & $t_\text{tr}$ (min) & $E$ (kJ) \\ 
                    \midrule
                    1 & \textbf{90.94} & \textbf{82.25} & 652 & 207 & {\color[HTML]{FF0000} \textbf{92.07}} & {\color[HTML]{FF0000} \textbf{83.45}} & 596 & {\color[HTML]{FF0000} \textbf{155}} \\
                    2 & 90.01 & 81.92 & \textbf{640} & \textbf{191} & 91.94 & 83.34 & {\color[HTML]{FF0000} \textbf{593}} & 157 \\
                    3 & 89.73 & 80.86 & 653 & 194 & 91.79 & 83.05 & 624 & 177 \\
                    4 & 89.90  & 81.08 & 845 & 223 & 91.65 & 82.75 & 645 & 173 \\
                    \bottomrule
            \end{tabular}  
            }

\end{wraptable}

\cref{tab: growth power} indicates that OSG is comparable or better than the other $m$-Shot Growth methods in all the metrics, thereby answering \textbf{Q2}. This result can be attributed to the lower training overhead of growth stage in OSG compared to the $m$-Shot Growth methods. It should be noted that as the growth steps increase, the accuracies go down and training cost goes up, thus indicating that the optimal solution cannot be found by further increasing the growth steps. Another comparison that is highlighted by~\cref{tab: growth power} is the one between GEARnn-1 and GEARnn-2. For each growth step, GEARnn-2 is better than the corresponding GEARnn-1 solution on all the metrics. The numbers highlighted in red indicate the best solution across the table. Thus,~\cref{tab: growth power} clearly highlights that 2-Phase approach using One-Shot Growth is the best combination to grow robust networks efficiently on the Edge. More comparisons between OSG and Multi-Shot growth are shown in~\cref{app: OSG vs Multi-Shot}.


\vspace{-10pt}
\section{Ablation Study}
\vspace{-10pt}
In this section, we look at the generalization of GEARnn to other robust augmentations and then understand the robustness and efficiency breakdowns for GEARnn.
\vspace{-5pt}

\vspace{-5pt}



\subsection{Generalization across robust augmentation methods}  
The results thus far employed AugMix~\cite{hendrycks2019augmix} tranforms ($\mathcal{T}$) to generate $\mathcal{D}_\text{aug}$ for robust training. In this section, we see if the benefits of GEARnn are maintained across other augmentation transforms. \cref{tab: PRIME} compares the implementation of PRIME~\cite{modas2022prime} augmentation across different methods. The accuracy and efficiency trend observed are similar to the results in~\cref{tab: GEARnn Quadro}. The important aspect to notice is the increase in training complexity gap ($\sim 2\times$) between GEARnn-1 and GEARnn-2. This is because OSG with PRIME is more expensive than OSG with AugMix.

\begin{minipage}{0.47\textwidth}    
        \setlength{\tabcolsep}{3pt}     
        \captionof{table}{Accuracy and Efficiency comparisons for PRIME ($\mathcal{D}_{\text{aug}}$) augmentation implemented for VGG-19 and CIFAR-10 on Quadro.}
        \label{tab: PRIME}          
        \centering
            \resizebox{\linewidth}{!}{
             \begin{tabular}{c c c c c}
                    \toprule
                    Method  & $\mathcal{A}_\text{cln}(\%)$ & $\mathcal{A}_\text{rob}$ (\%)& $t_\text{tr}$ (min) & $E$ (kJ)\\
                    \midrule
                    Small ($\mathcal{D}_{\text{in}}$) & 92.69 & 70.57\textcolor{red}{$\downarrow$} & 31 & 241\\
                    \midrule
                    Small ($\mathcal{D}_{\text{aug}}$) &  \textbf{\underline{91.30}} & \textbf{\underline{87.01}} & 829 & 2550\\
                    GEARnn-1 &  88.37 & 83.18 & \textbf{458} & \textbf{1410}\\
                    GEARnn-2 &  \textbf{90.26} & \textbf{84.45} & \textbf{\underline{234}} & \textbf{\underline{856}}\\
                    \bottomrule
            \end{tabular}
            }
\end{minipage}
\hfill
\begin{minipage}{0.47\textwidth}    
        \setlength{\tabcolsep}{3pt}     
        \captionof{table}{Training time and energy breakdown for GEARnn on CIFAR-10 using VGG-19 on Quadro.}
        \label{tab: GEARnn energy breakdown}          
        \centering
            \resizebox{\linewidth}{!}{
             \begin{tabular}{c | c c c | c c c c}
                    \toprule
                    \multirow{2}{*}{Quantity} & \multicolumn{3}{|c|}{GEARnn-1} & \multicolumn{4}{c}{GEARnn-2} \\
                    \cmidrule{2-8}
                    & OSG-1 & OSG-2 & Total & OSG-1 & OSG-2 & ERA & Total\\
                    \midrule
                    training & 38 & 48 & 86 & 5 & 10 & 38 & 53\\
                    time (min) & 44\% & 56\% & 100\% & 9\% & 19\% & 72\% & 100\%\\
                    \midrule
                    energy & 180 & 372 & 552 & 26 & 71 & 201 & 298\\
                    (kJ) & 33\% & 67\% & 100\% & 9\% & 24\% & 67\% & 100\%\\
                    \bottomrule
            \end{tabular}
            }
\end{minipage}
\vspace{-5pt}
\subsection{Efficiency and Robustness breakdown}
\vspace{-5pt}
\begin{wraptable}{}{0.45\columnwidth}
        \centering
        \vspace{-5pt}

        \setlength{\tabcolsep}{3pt}     
        \caption{Impact of using OSG and ERA for CIFAR-100 and VGG-19 on Quadro.}
        \label{tab: GEARnn robustness ablation}          
        \centering
            \resizebox{\linewidth}{!}{
             \begin{tabular}{c c | c c | c c c}
                    \toprule
                    \multicolumn{2}{c|}{Phase-1 ($\mathcal{D}_{\text{in}}$)} & \multicolumn{2}{c|}{Phase-2 ($\mathcal{D}_{\text{aug}}$)} & \multirow{2}{*}{$\mathcal{A}_{\text{rob}}(\%)$} & \multirow{2}{*}{$t_\text{tr}$(min)} & \multirow{2}{*}{$E$ (kJ)}\\
                    \cmidrule{1-4}
                    vanilla & OSG & AugMix & ERA & & &\\
                    \midrule
                    \checkmark &  &  &  & 38.72 & 18 & 161\\
                    & \checkmark &  &  & 38.01 & 16 & 118\\
                    &  &  \checkmark &  & 46.50 & 62 & 385\\
                    &  &  & \checkmark & 46.13 & 46 & 406\\
                    \checkmark &  & \checkmark &  & 53.74 & 79 & 534 \\
                    & \checkmark &  & \checkmark & \textbf{54.31} & 64 & 515\\
                    \bottomrule
            \end{tabular}
            \vspace{-5pt}
            }    
        
\end{wraptable}
\cref{tab: GEARnn energy breakdown} shows the breakdown of energy and training time for different stages of GEARnn-1 and GEARnn-2. OSG-1 involves the training of backbone $f_0$ and OSG-2 includes both the growth stage and training of $f_g$. The key aspect to notice in \cref{tab: GEARnn energy breakdown} is the small fraction of training cost required by OSG-1 and OSG-2 in GEARnn-2 to provide a good initialization.

\cref{tab: GEARnn robustness ablation} shows the ablation studies of different components used in GEARnn-2 and compares it with a fixed network robust training. Firstly, we notice that OSG is more efficient than vanilla (fixed network) training, both in terms of training time and energy while achieving comparable accuracy. Similar observations can be made for ERA over AugMix. Performing 2-Phase approach by using either vanilla or OSG as initialization provides a significant boost in robustness while incurring a minimal overhead in training cost. Thus the 2-Phase approach is a clear winner over the 1-Phase approach, and in particular the combination of OSG and ERA used for GEARnn-2 is optimal. More comparisons between AugMix and ERA on Jetson are shown in~\cref{app: ERA on Jetson}.
\vspace{-10pt}

\section{Discussion}
\label{sec: Discussion}
\vspace{-10pt}
Until now we have looked at extensive empirical simulations that highlight the efficacy of GEARnn-2. In this section, we will look at the inner workings of this algorithm. Specifically, we will see what network topologies are generated when OSG designs compact networks, and also understand why clean data initialization benefits robust training.
\vspace{-10pt}
\subsection{Impact of OSG on Network Topology}
\label{subsec: understand growth}
\vspace{-5pt}
\begin{figure*}[h]
\centering
\begin{subfigure}{.3\textwidth}
  \centering
  \includegraphics[width=\linewidth]{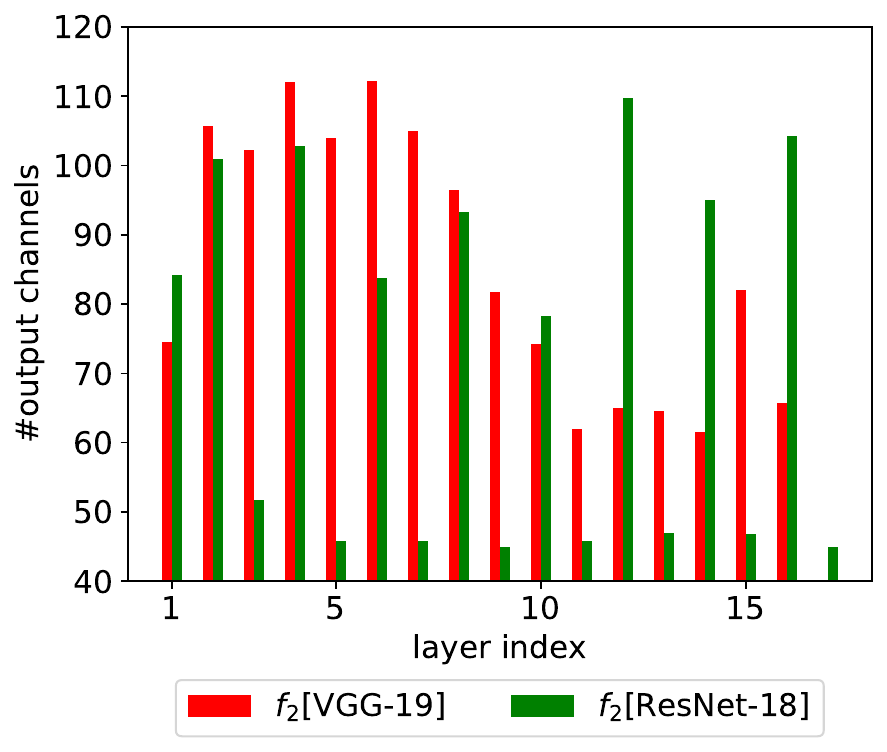}
  \subcaption{\label{fig: resmob topos}}
\end{subfigure}
\begin{subfigure}{.31\textwidth}
  \centering
  \includegraphics[width=\linewidth]{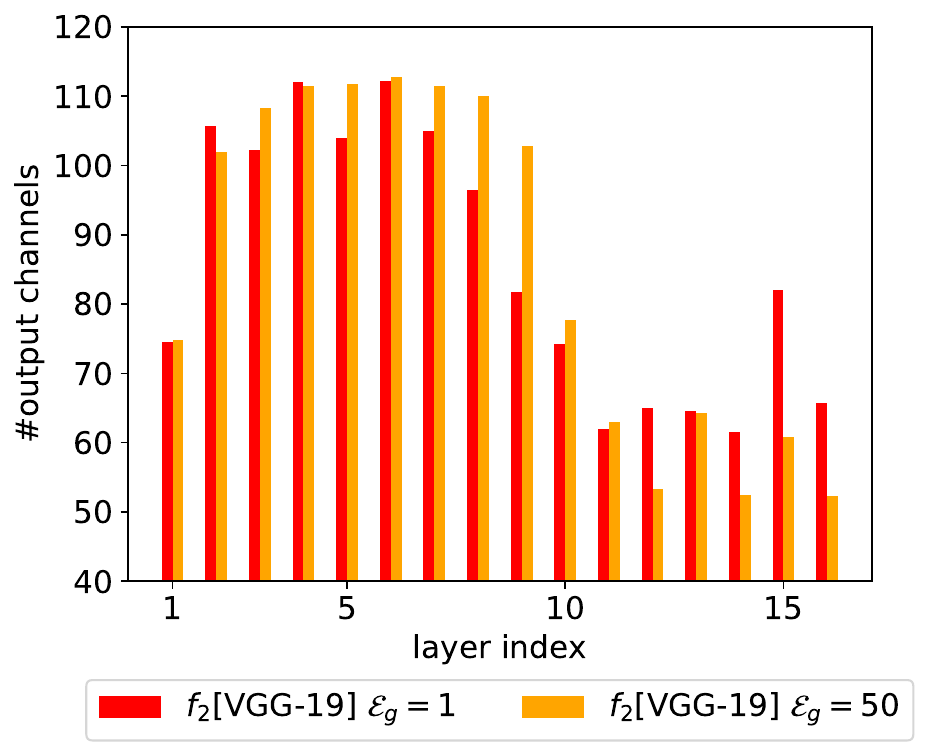}
  \subcaption{\label{50 growth epochs}}
\end{subfigure}
\begin{subfigure}{.33\textwidth}
  \centering
  \includegraphics[width=\linewidth]{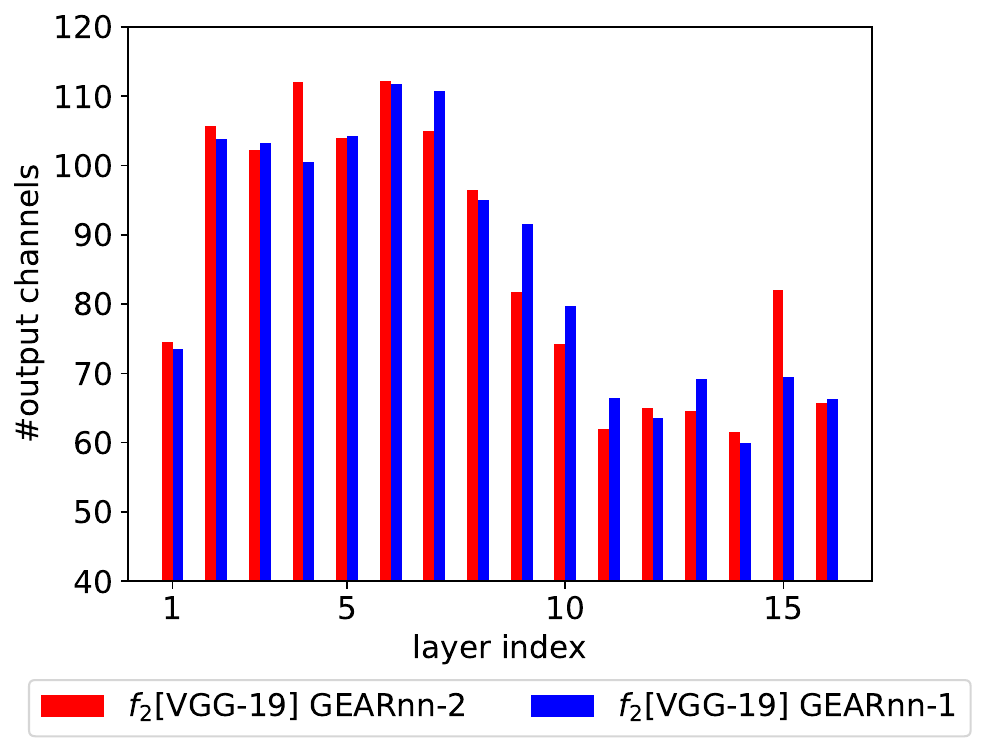}
  \subcaption{\label{fig: robusttopo}}
\end{subfigure}
\caption{\label{fig: growth understanding} Average output channels vs. layer index for CIFAR-10 on Quadro is shown. Plot (a) looks at the impact of network architecture and highlights the non-uniform growth pattern in plain CNNs versus steady zigzag pattern in residual CNNs. Plots (b) and (c) indicate that modifying the number of growth epochs ($\mathcal{E}_g$) or performing 1-Phase robust growth does not affect the topology pattern much. 
\vspace{-10pt}}
\end{figure*}
In this section, we look at the growth topology patterns ($\{w_l\}_{l=1}^L$) as a function of layer index $l$. Specifically, we investigate these patterns in the simple setting of OSG ($\mathcal{D}_{\text{in}}$) implemented on CIFAR-10 for ($\mathcal{E}_1, \mathcal{E}_2$) = (40, 40) and an initial backbone $f_0$ with $\{w_l\}_{l=1}^L = 45$. The bar plots represent the mean width ($\mathbb{E}[w_l]$) across four random seeds.

\textbf{Backbone architecture: }For plain CNNs like VGG-19~\cite{simonyan2014very} - the initial layers have higher number of convolutional filters compared to final layers. This correlates with the observations seen in quantization~\cite{sakr2018analytical} where the initial layers require higher precision compared to the final layers. 
However, in case of residual networks like ResNet-18, the pattern is largely invariant to network depth and is oscillating as shown in~\cref{fig: resmob topos}. The invariance in depth can be attributed to the direct gradient flow facilitated by the shortcut connections making each residual block act independently of the depth.
In each residual block, the macro-level pattern in plain CNNs is observed at a micro-level, i.e. initial layer has more output channels than the final layer.

\textbf{Growth Epochs and Data: }All the above experiments were performed for a single growth epoch ($\mathcal{E}_g$~=~1) and on clean data. The effect of increasing $\mathcal{E}_g$ to 50 and using ERA data for growth (GEARnn-1) is shown in \cref{50 growth epochs} and \cref{fig: robusttopo}. The topology pattern in both cases remains roughly the same as OSG ($\mathcal{D}_{\text{in}}$) $\mathcal{E}_g=1$.
\vspace{-10pt}
\subsection{Rationale for 2-Phase Approach}
\begin{figure}[h]
    \vspace{-10pt}
    \begin{subfigure}{.47\textwidth}
        \centering
        \includegraphics[width=\linewidth]{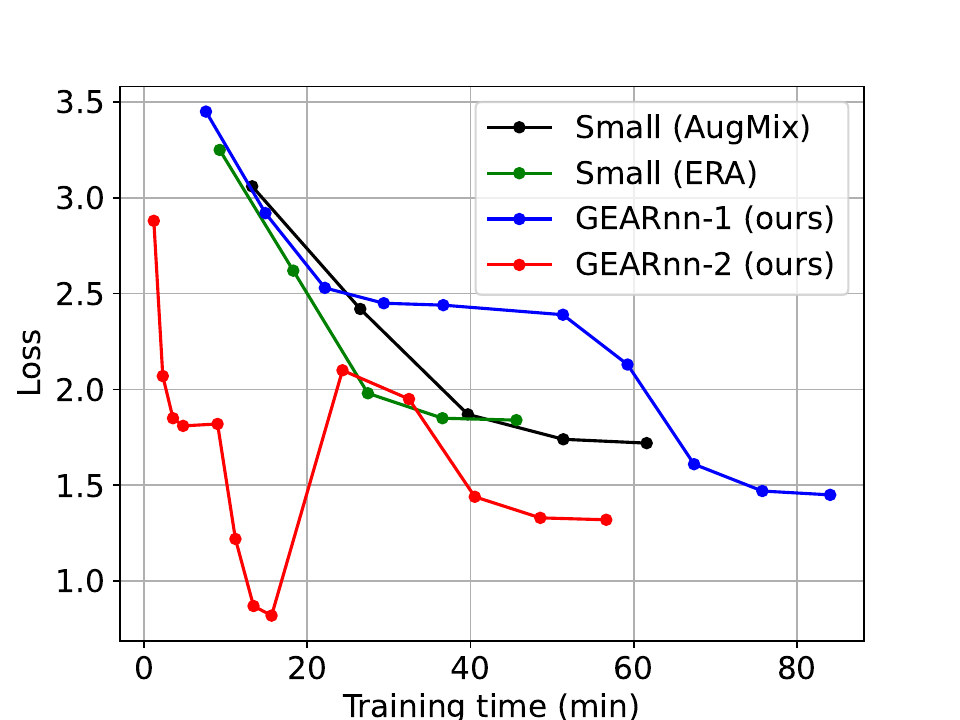}
        \subcaption{\label{fig: 1D loss curve 1}}
    \end{subfigure}
    \begin{subfigure}{.47\textwidth}
        \centering
        \includegraphics[width=\linewidth]{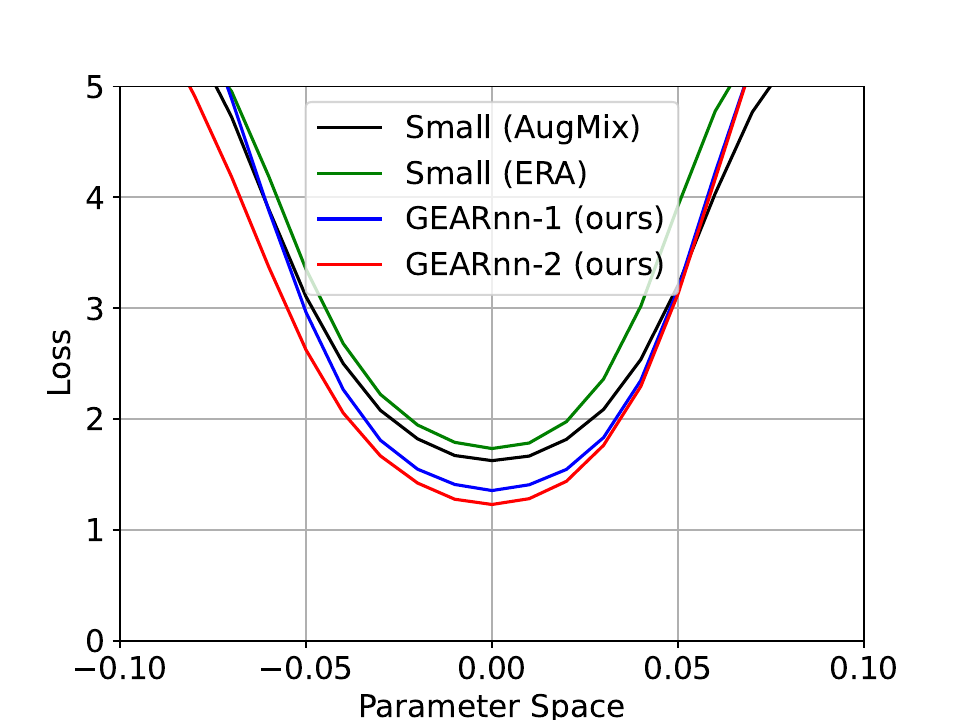}
        \subcaption{\label{fig: 1D loss curve 2}}
    \end{subfigure}
    \vspace{-5pt}
    \caption{\label{fig: loss curves} Loss comparisons for 2-Phase (GEARnn-2) and 1-Phase (rest) approaches for CIFAR-100 and VGG-19 on Quadro with 50 epochs of robust training at final model size.~\cref{fig: 1D loss curve 1} highlights that GEARnn-2 loss converges to the minimum faster than other approaches.~\cref{fig: 1D loss curve 2} shows the loss landscapes where GEARnn-2 achieves the smallest minima with a wide curve, thus aiding better generalization~\cite{li2018visualizing}.}
\end{figure}

\vspace{-5pt}
In this section, we provide insights for the efficacy of GEARnn-2 and the 2-Phase approach. In particular, we highlight why training or growth done on clean data provides a good initialization for robust training. We look at the loss curves for the 1-Phase approaches (Small (AugMix), Small (ERA), GEARnn-1) and the 2-Phase approach (GEARnn-2) in~\cref{fig: loss curves}. The initial dip in GEARnn-2 loss function in~\cref{fig: 1D loss curve 1} is due to the loss landscape being different for Phase-1 done on clean data compared to Phase-2 done on augmented data. One can clearly see that GEARnn-2 achieves a lower loss at a faster rate compared to the other 1-Phase approaches, thus justifying the importance of clean growth initialization. We also plot the filter normalized loss curves~\cite{li2018visualizing} in~\cref{fig: 1D loss curve 2} to observe the loss landscapes around the converged weights. GEARnn-2 finds the smallest minima while also having a wide curve which enables better generalization~\cite{li2018visualizing}.

\begin{figure*}[h]
\centering
\begin{subfigure}{.19\textwidth}
  \centering
  \includegraphics[width=\linewidth]{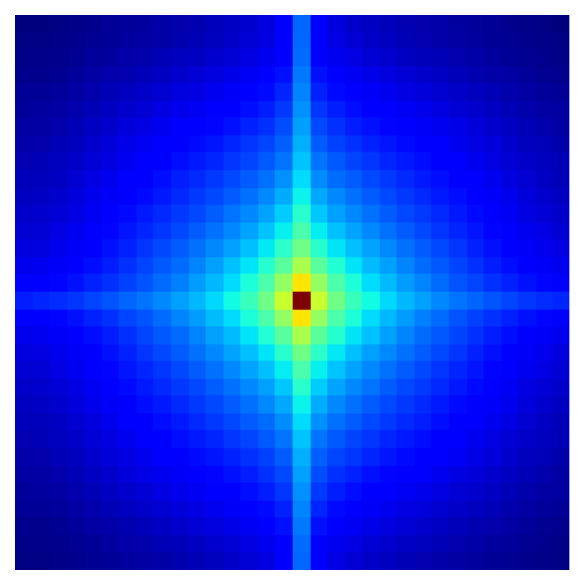}
  \subcaption{\label{fig: Clean Image} clean data}
\end{subfigure}
\begin{subfigure}{.19\textwidth}
  \centering
  \includegraphics[width=\linewidth]{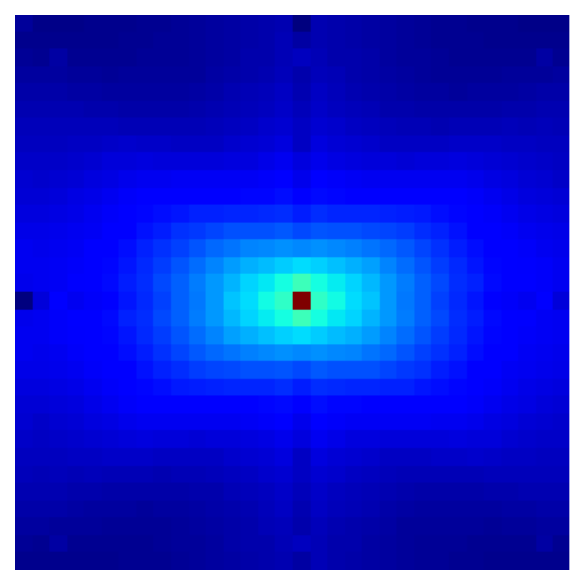}
  \subcaption{\label{fig: snow} snow}
\end{subfigure}
\begin{subfigure}{.19\textwidth}
  \centering
  \includegraphics[width=\linewidth]{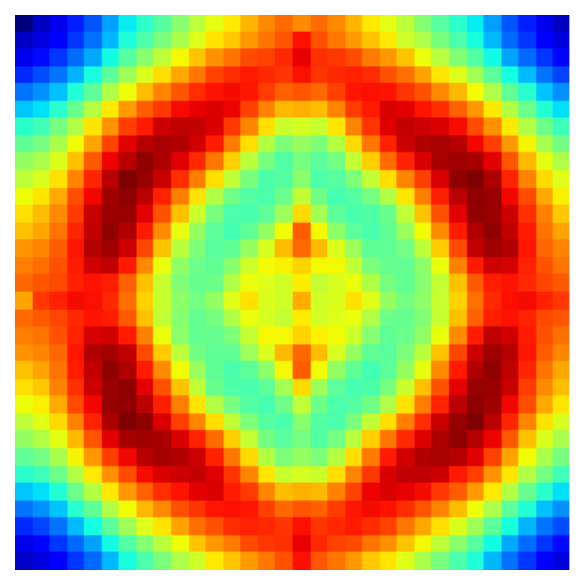}
  \subcaption{\label{fig: JPEG compression} JPEG comp.}
\end{subfigure}
\begin{subfigure}{.19\textwidth}
  \centering
  \includegraphics[width=\linewidth]{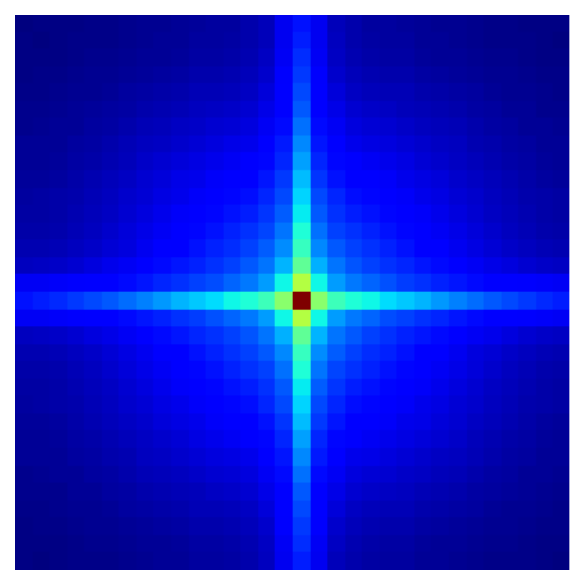}
  \subcaption{\label{fig: all augmentations} AugMix}
\end{subfigure}
\begin{subfigure}{.19\textwidth}
  \centering
  \includegraphics[width=\linewidth]{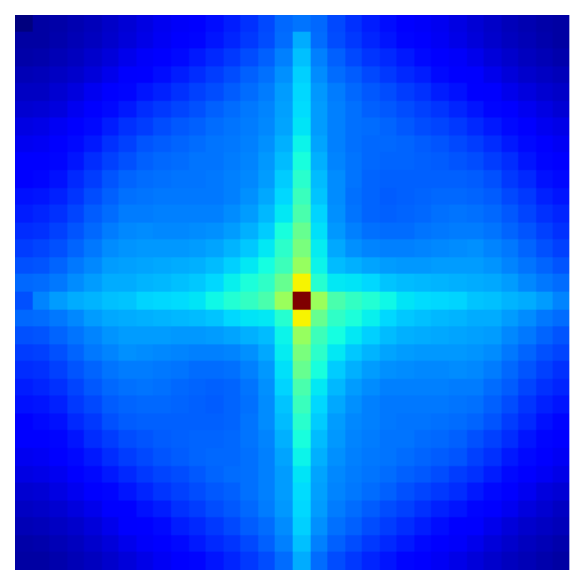}
  \subcaption{\label{fig: Mean Corruption} all corruptions}
\end{subfigure}
\caption{\label{fig: Fourier Spectrum} The Fourier spectrum of clean images (from CIFAR-10), their corresponding augmented (AugMix) and corrupted versions (CIFAR-10-C at severity 3) are shown. The augmentation and corruption spectrums (Figs.~\ref{fig: snow},~\ref{fig: JPEG compression}, ~\ref{fig: all augmentations} \&~\ref{fig: Mean Corruption}) are obtained by taking Fourier Transform of the difference with the clean image (Eg: $\kappa(\mathbf{x_{\text{in}}},3)-\mathbf{x_{\text{in}}}$). Snow(\cref{fig: snow}) and JPEG compression(\cref{fig: JPEG compression}) corruptions are shown to highlight the range of possible frequencies in the corrupted spectrums. The similarity in the spectrums of clean (\cref{fig: Clean Image}), augmented (\cref{fig: all augmentations}) and all-corrupted (\cref{fig: Mean Corruption}) images highlights the importance of OSG initialization using clean data.
\vspace{-5pt}}
\end{figure*}

The above explanation illustrates why GEARnn-2 has a good training and generalization performance. However, in order to understand why initialization with \textit{clean data} aids faster convergence of robust training, we look at the Fourier spectrums of the clean, augmented and corrupted images in~\cref{fig: Fourier Spectrum}.~\cref{fig: Clean Image} indicates that the clean images lie in the low frequency domain, while the corrupted samples occupy a wide range of frequencies (Figs.~\ref{fig: snow} \&~\ref{fig: JPEG compression}). Crucially, the spectrum containing all the augmentations (in AugMix)~\cref{fig: all augmentations} and all the corruptions (in CIFAR-10-C)~\cref{fig: Mean Corruption} is also in the low-frequency domain, similar to the clean image spectrum~\cref{fig: Clean Image}. This is unlike the scenario of adversarial or Gaussian noise perturbations, which lie in the high-frequency domain~\cite{yin2019fourier} and hence may not benefit from clean data initialization. Thus, robust training for common corruptions benefits from initialization with clean data.

\vspace{-10pt}
\section{Limitations and Broader Impacts}
\label{subsec: Limitations}
\vspace{-10pt}
While our work has conclusively shown that a 2-Phase approach for growing robust networks is computationally efficient, a theoretical convergence analysis for this result is currently lacking. Such a result would help identify favorable initial conditions for robust training to achieve high accuracy in fewer epochs. 

The impact of our work is broadly positive since it enables efficient robust training on Edge devices. We do not see any direct negative impact of our work.
\vspace{-10pt}
\section{Conclusion}
\vspace{-10pt}
\label{sec:conclusion}
We addressed the problem of growing robust networks efficiently on Edge devices. Specifically, we concluded that a 2-Phase approach with distinct clean growth and robust training phases is significantly more efficient than a 1-Phase approach which employs augmented data for growth. We encapsulated this result into the GEARnn algorithm and experimentally demonstrated its benefits on a real-life Edge device. An interesting and non-trivial extension of our work would be to use unlabeled data for growing efficient and robust networks. Another extension would be to design robust networks for complex tasks such as object detection on highly resource-constrained Edge platforms.
\vspace{-10pt}
\section{Reproducibility Statement}
\vspace{-10pt}
We list all the experimental setup details in~\cref{sec: setup} and~\cref{app: train setup}. We use fixed seeds during the simulation runs so that our results can be reproduced. We will make the code public along with the software versions if the paper is accepted so that the community can use and reproduce our results.
\clearpage
\bibliography{iclr2025_conference}
\bibliographystyle{iclr2025_conference}

\newpage
\appendix
\section*{Appendix / Supplemental material}
\vspace{-5pt}
\vspace{-5pt}
\section{Training Setup}
\label{app: train setup}
\vspace{-5pt}
{\bf Hyperparameters:} The setup for growth and robust augmentation follows closely with what is described in Firefly~\cite{wu2020firefly} and AugMix~\cite{hendrycks2019augmix}, respectively. The parametric training is done for 160 epochs using a batch-size of 128 and an initial learning rate of 0.1. The learning rate scheduler decays by 0.1 at half and three-fourths of the total number of epochs. We use the Swish loss function for MobileNet-V1 as used in ~\cite{wu2020firefly}, while employing ReLU for the other two networks. Instead of using three fully-connected layers at the end of VGG-19, we use only one as done in~\cite{wu2020firefly}. Stochastic Gradient Descent (SGD) optimizer is used with momentum 0.9 and weight decay $10^{-4}$. As for the standard growth process, we use a Root Mean Square Propagation (RMSprop) optimizer with momentum 0.9, alpha 0.1 and initial learning rate of $9\times10^{-5}$. The number of workers is chosen as 4. For ERA, $(W,D,J) = (1,3,4)$ is picked. The augmentation transforms $\mathcal{T}$ are same as that of AugMix~\cite{hendrycks2019augmix} for all the results except~\cref{tab: PRIME}, where we pick the transforms from PRIME~\cite{modas2022prime}. As specified in AugMix, we also do not use any augmentations which are directly present in the corrupted test dataset.

In case of OSG, the initial backbone $f_0$  is chosen as a network with $w_l = 45$ for all $l = \{1, ..., L\}$ and is thus extremely small. 
The number of randomly initialized neurons at each growth stage is 70. We ensure that $\mathcal{E}_2$ of GEARnn-1 and $\mathcal{E}_r$ of GEARnn-2 are same for a fair comparison. $\mathcal{E}_g$ is chosen as 1 based on Firefly~\cite{wu2020firefly}.

\textbf{Jetson Training:} The two changes to the GEARnn algorithm when implementing on NVIDIA Jetson Xavier are - one we use $j=3$ instead of $j=4$, and two, we allow only 40 randomly initialized new neurons per layer in the growth step (as compared to 70 in~\cite{wu2020firefly}). These measures are taken to stay within the memory constraints of the Edge device. We also reduce the batch size (and learning rate) appropriately in case the above measures are insufficient.

\section{Ablation Studies}
\label{app: Ablation Studies}
\subsection{Diagnostics of Robust Augmentation Methods}
\label{app: ERA diagnosis}
\begin{minipage}{\textwidth}
  \begin{minipage}[b]{0.43\textwidth}
        In this section, we investigate which aspects of the robust augmentation framework described in~\cref{subsec: efficient AugMix} contribute most to the robustness while being training efficient. \cref{tab: AugMix modifcations} shows different modifications of the stochastic chains obtained by varying $(W,D,J)$ values. It can be observed that the basic version with $(W,D,J) = (1,1,0)$ (uses only standard cross entropy loss with the label and augmented data as input) has the least training time, but suffers a significant drop in $\mathcal{A}_{\text{rob}}$ compared to standard AugMix. Crucially, we note that increase in $D$ and $J$ has more impact on robustness at a lesser training cost compared to $W$. For ERA, we pick the modification with $(W,D,J) = (1,3,4)$ as it provides the highest robustness while simultaneously reducing training time over AugMix.

  \end{minipage}
  \hfill
  \begin{minipage}[b]{0.54\textwidth}
    \centering
    \centering
        \resizebox{\columnwidth}{!}{
            \begin{tabular}{c c c c c c }
                \toprule
                    {Experiment}   & {$W$} & {$D$} & {$J$} & $\mathcal{A}_{\text{rob}}(\%)$ & $t_\text{tr}$(min)\\
                    \midrule
                    \midrule
                    Basic   & 1& 1& 0& 77.74& \textbf{10}\\
                    + width& 3& 1& 0& 78.51& 16\\
                    + depth      & 1& 3& 0& 80.31& 12\\
                    +  JSD-3       & 1& 1& 3& 82.43& 20\\
                    + width + depth& 3& 3& 0& 80.47& 21\\
      + width + JSD-3& 3& 1& 3& 82.27& 32\\
                    + depth + JSD-3& 1& 3& 3& 83.67& 22\\
                    + depth + JSD-2& 1  & 3& 2& 82.41& 13\\
                    \textbf{+ depth + JSD-4}& 1  & 3& 4& \textbf{84.10} & 29\\
                    AugMix~\cite{hendrycks2019augmix}  & 3  & 3& 3& 84.05& 41\\
                    \bottomrule
                    \end{tabular}
            }
    \captionof{table}{\label{tab: AugMix modifcations}Impact of training AugMix-variants on the robust accuracy and training time. Network $f_2$ from OSG is used as the starting network and $\mathcal{E}_r$ = 40. All the methods are implemented for CIFAR-10 and 5\% VGG-19 network on Quadro. $W, D, J$ represent the width, depth and consistency samples used in the stochastic chains.}
    \end{minipage}
  \end{minipage}
\clearpage
\subsection{Benefits of ERA on Jetson}
\label{app: ERA on Jetson}
\begin{wraptable}{}{0.45\columnwidth}
    
        \vspace{-15pt}
        \setlength{\tabcolsep}{3pt}     
        \caption{Comparison of ERA versus AugMix on Jetson for VGG-19 and CIFAR-10 when trained for 160 epochs
        }
        \label{tab: ERA vs AugMix on Jetson}          
        \centering
            \resizebox{0.9\linewidth}{!}{
             \begin{tabular}{c | c c c c }
                    \toprule
                    Method & $\mathcal{A}_\text{cln}(\%)$ & $t_\text{tr}$ (min) & $E$ (kJ)\\ 
                    \midrule
                    Small (AugMix)  & 93.36 & 85.73 & 1543 & 522\\
                    Small (ERA)  & \textbf{93.42} & \textbf{85.74} & \textbf{1542} & \textbf{486}\\
                    \bottomrule
            \end{tabular}
            \vspace{-5pt}  
            }
\end{wraptable}
In this section we will look at the benefits of using ERA over AugMix. Previously, we had looked at this comparison on Quadro using CIFAR-100 and VGG-19 in~\cref{tab: GEARnn robustness ablation}. Here, we will look at these results for CIFAR-10 and VGG-19 on Jetson when training a fixed-size Small ($\mathcal{D}_{\text{aug}}$) network for 160 epochs.~\cref{tab: ERA vs AugMix on Jetson} indicates that ERA is better than AugMix on all the metrics. Thus our choice of ERA over AugMix is justified.
  
\subsection{OSG versus Multi-Shot Growth Comparisons}
\label{app: OSG vs Multi-Shot}
In this section, we first look at clean data growth comparisons on Jetson in~\cref{tab: clean OSG vs Small}. Then we look at robust data growth comparisons on Quadro in~\cref{tab: GEARnn1 multi-shot}. When comparing various growth methods on clean data in~\cref{tab: clean OSG vs Small}, we also include the Small ($\mathcal{D}_{\text{in}}$) results to highlight the efficiency benefits of growth. We can see that OSG has comparable or better training efficiency than all the methods including Small ($\mathcal{D}_{\text{in}}$). In case of clean accuracy, we observe that OSG has the highest among growth methods while being slightly lower than Small ($\mathcal{D}_{\text{in}}$). Looking at~\cref{tab: GEARnn1 multi-shot}, we see that for both datasets OSG again provides comparable or best solution among all the growth methods across all metrics. Thus our choice of OSG over other Multi-Shot Growth methods is justified.
\begin{table}[!hbt]
    \parbox{.45\linewidth}{
        \setlength{\tabcolsep}{3pt}     
        \caption{Comparison of training complexities and clean accuracy for different growth methods implemented using VGG-19 and clean CIFAR-10 data on Jetson.}
        \label{tab: clean OSG vs Small}          
        \centering
            \resizebox{0.75\linewidth}{!}{
             \begin{tabular}{c | c c c }
                    \toprule
                    Growth Steps & $\mathcal{A}_\text{cln}(\%)$ & $t_\text{tr}$ (min) & $E$ (kJ)\\ 
                    \midrule
                    Small ($\mathcal{D}_{\text{in}}$)  & 91.96 & 267 & 73\\
                    \midrule 
                    1 & \textbf{90.80} & 210 & \textbf{51} \\
                    2 & 90.49& \textbf{209} & 53 \\
                    3 & 90.31&  231 & 59 \\
                    4 & 90.08 & 275 & 65 \\
                    \bottomrule
            \end{tabular}
            \vspace{-5pt}
        }
    }
\hfill
    \parbox{.52\linewidth}{
        \setlength{\tabcolsep}{3pt}     
        \caption{OSG versus Multi-Shot Growth using ERA data, i.e. GEARnn-1 with Multi-Shot Growth. Results are shown for VGG-19 on Quadro.}
        \label{tab: GEARnn1 multi-shot}          
        \centering
            \resizebox{\linewidth}{!}{
             \begin{tabular}{c | c c c | c c c}
                    \toprule
                    Growth & \multicolumn{3}{|c|}{CIFAR-10} & \multicolumn{3}{c}{CIFAR-100} \\
                    \cmidrule{2-7}
                    Steps & $\mathcal{A}_\text{rob}$ (\%)& $t_\text{tr}$ (min) & $E$ (kJ) & $\mathcal{A}_\text{rob}$ (\%)& $t_\text{tr}$ (min) & $E$ (kJ)\\
                    \midrule
                    1 & \textbf{82.86} & \textbf{75} & 449 & \textbf{52.68} & \textbf{84} & \textbf{426} \\
                    2 & 82.31 & 81 & \textbf{329} & 51.34 & 100 & 554 \\
                    3 & 81.94 & 82 & 506 & 50.52 & 103 & 517 \\
                    4 & 81.81 & 86 & 389 & 50.45 & 102 & 590 \\
                    \bottomrule
            \end{tabular}
            \vspace{-5pt}
        }
    }
\end{table}

\clearpage
\section{Impact of Robust Training Epochs}
\label{app: robust epochs}
\begin{figure*}[h]
\centering
\begin{subfigure}{.32\textwidth}
  \centering
  \includegraphics[width=\linewidth]{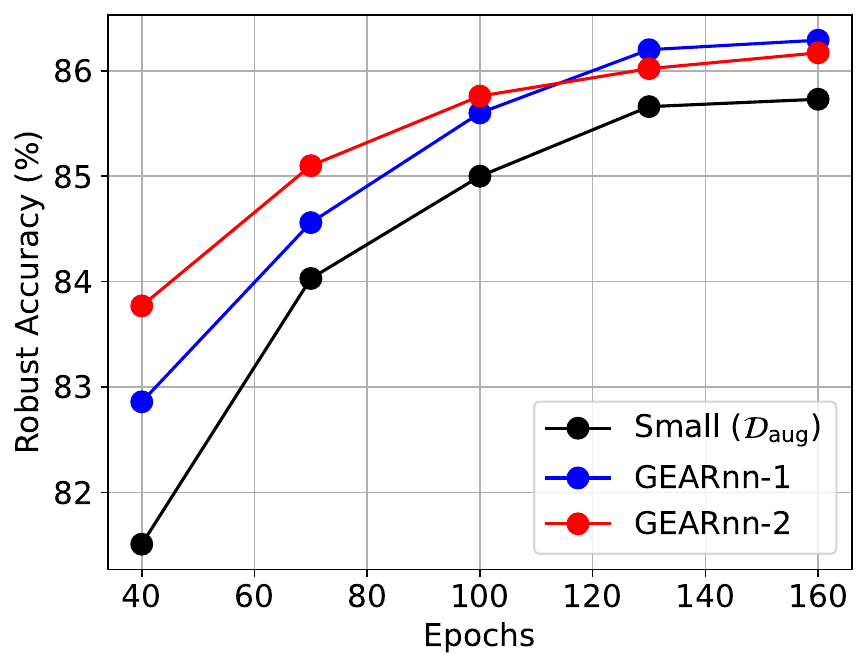}
  \subcaption{\label{c10 clean}}
\end{subfigure}
\begin{subfigure}{.32\textwidth}
  \centering
  \includegraphics[width=\linewidth]{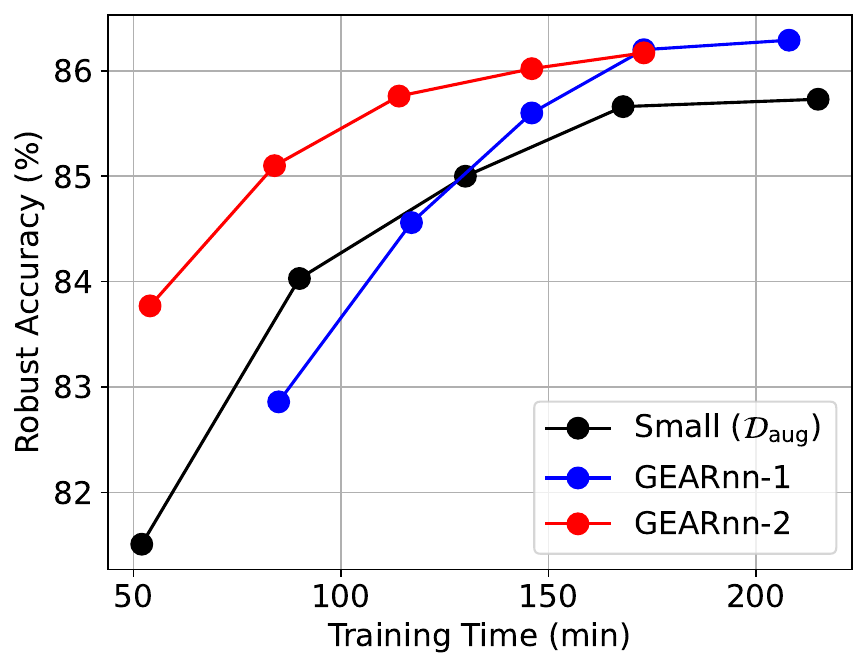}
  \subcaption{\label{c10 robustness}}
\end{subfigure}
\begin{subfigure}{.32\textwidth}
  \centering
  \includegraphics[width=\linewidth]{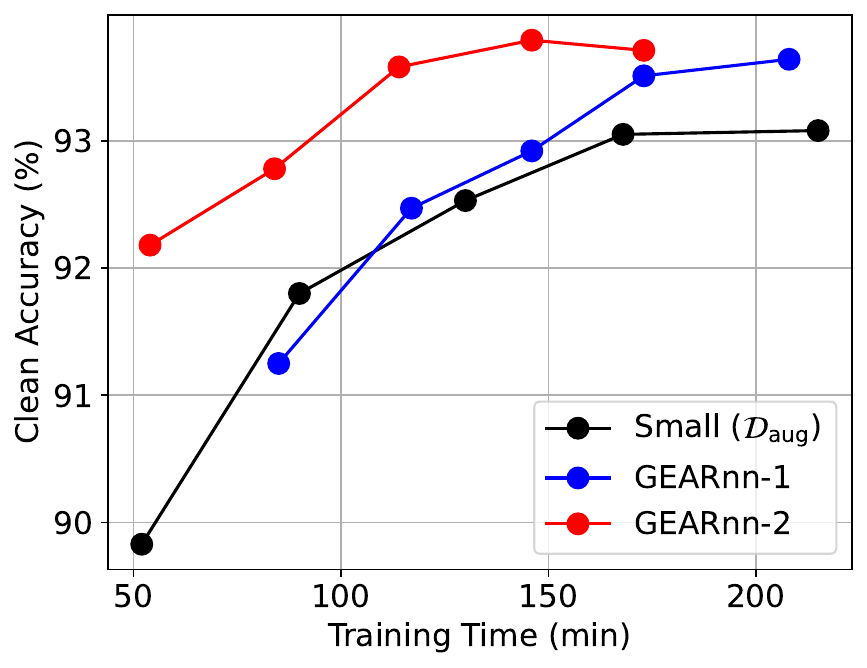}
  \subcaption{\label{c10 traintime}}
\end{subfigure}

\begin{subfigure}{.32\textwidth}
  \centering
  \includegraphics[width=\linewidth]{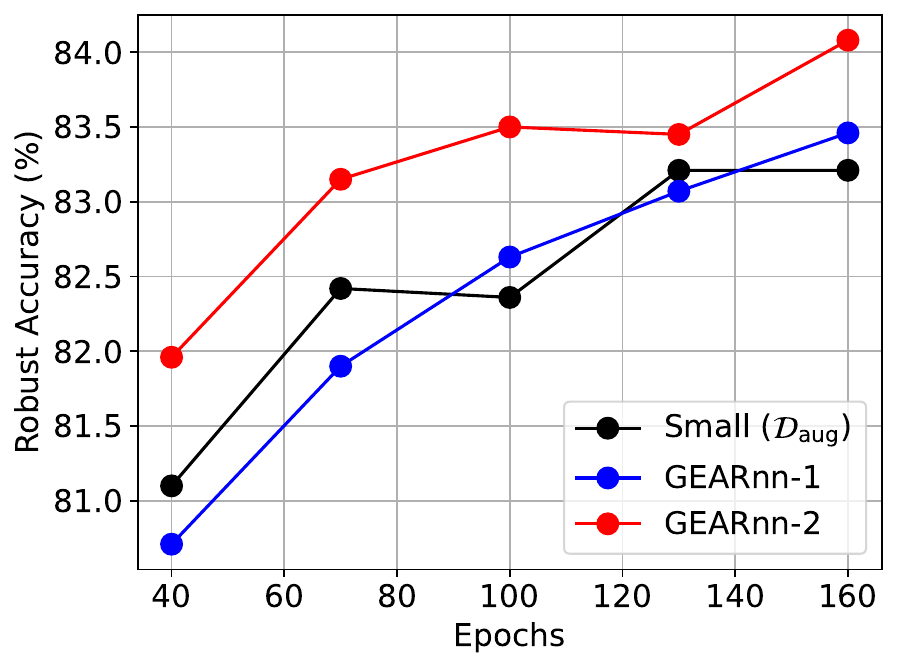}
  \subcaption{\label{c100 clean}}
\end{subfigure}
\begin{subfigure}{.32\textwidth}
  \centering
  \includegraphics[width=\linewidth]{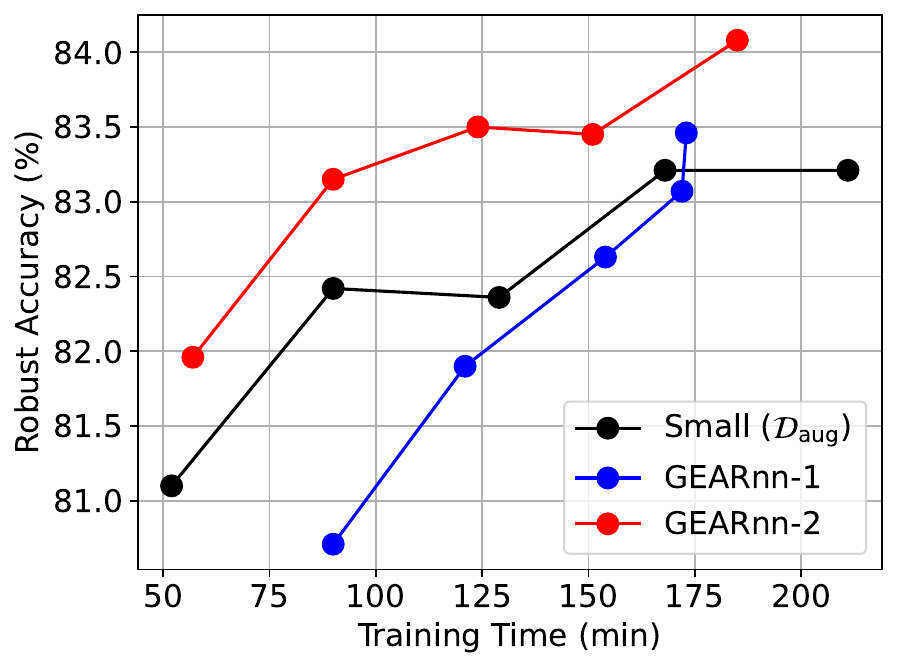}
  \subcaption{\label{c100 robustness}}
\end{subfigure}
\begin{subfigure}{.32\textwidth}
  \centering
  \includegraphics[width=\linewidth]{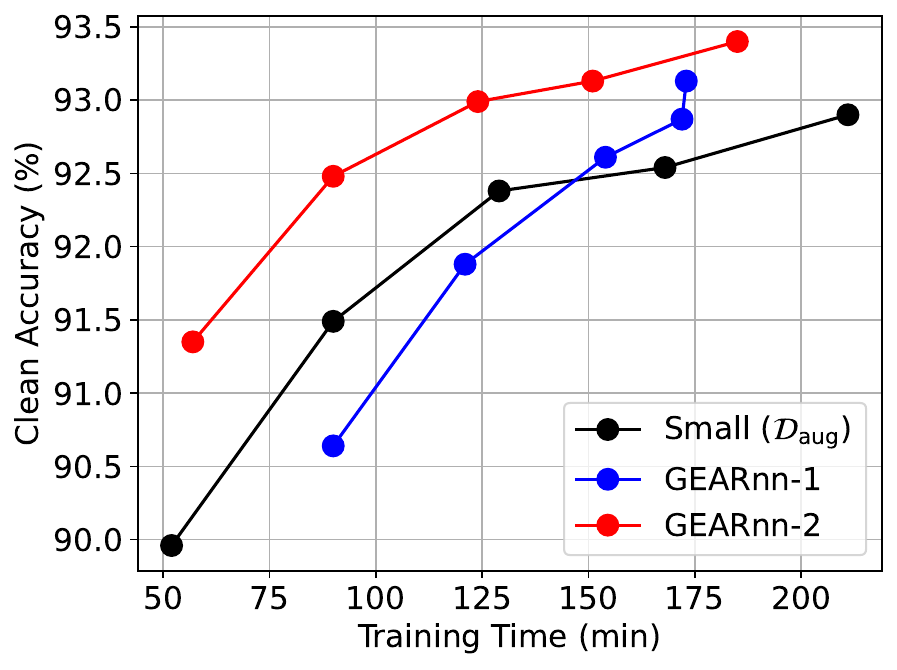}
  \subcaption{\label{c100 traintime}}
\end{subfigure}

\begin{subfigure}{.32\textwidth}
  \centering
  \includegraphics[width=\linewidth]{figures/arob_epoch.pdf}
  \subcaption{\label{c10 clean}}
\end{subfigure}
\begin{subfigure}{.32\textwidth}
  \centering
  \includegraphics[width=\linewidth]{figures/arob_ttr.pdf}
  \subcaption{\label{c10 robustness}}
\end{subfigure}
\begin{subfigure}{.32\textwidth}
  \centering
  \includegraphics[width=\linewidth]{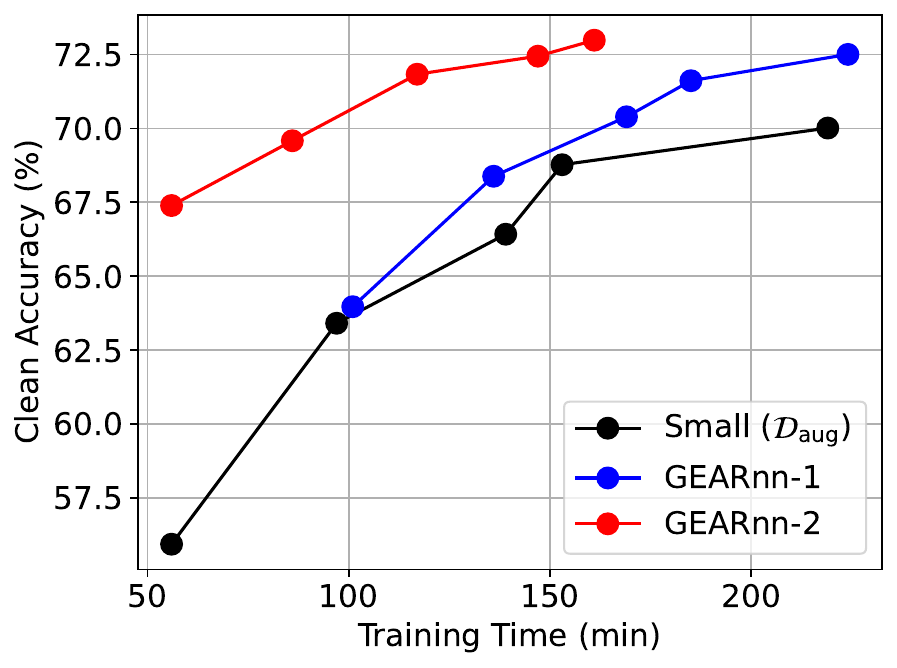}
  \subcaption{\label{c10 traintime}}
\end{subfigure}

\begin{subfigure}{.32\textwidth}
  \centering
  \includegraphics[width=\linewidth]{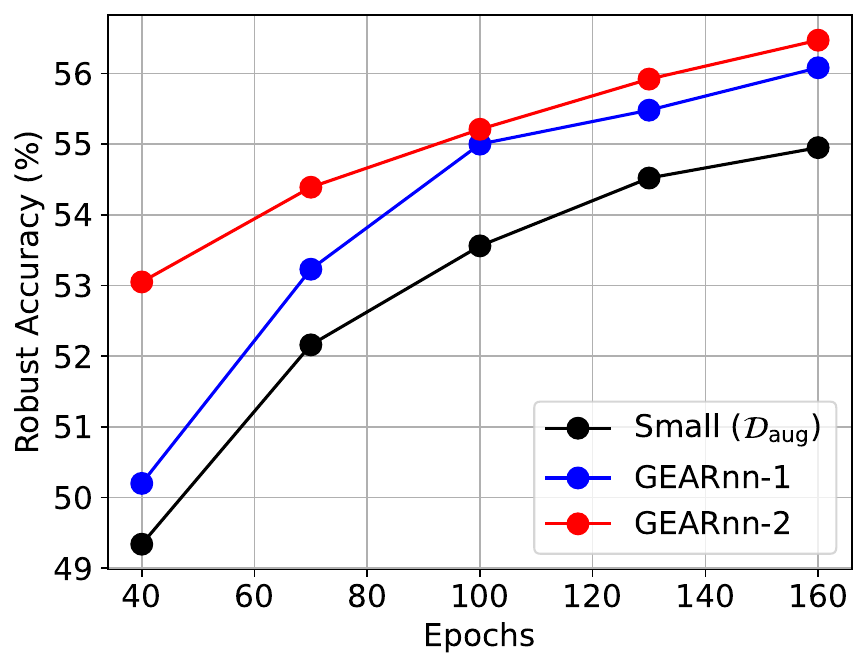}
  \subcaption{\label{c100 clean}}
\end{subfigure}
\begin{subfigure}{.32\textwidth}
  \centering
  \includegraphics[width=\linewidth]{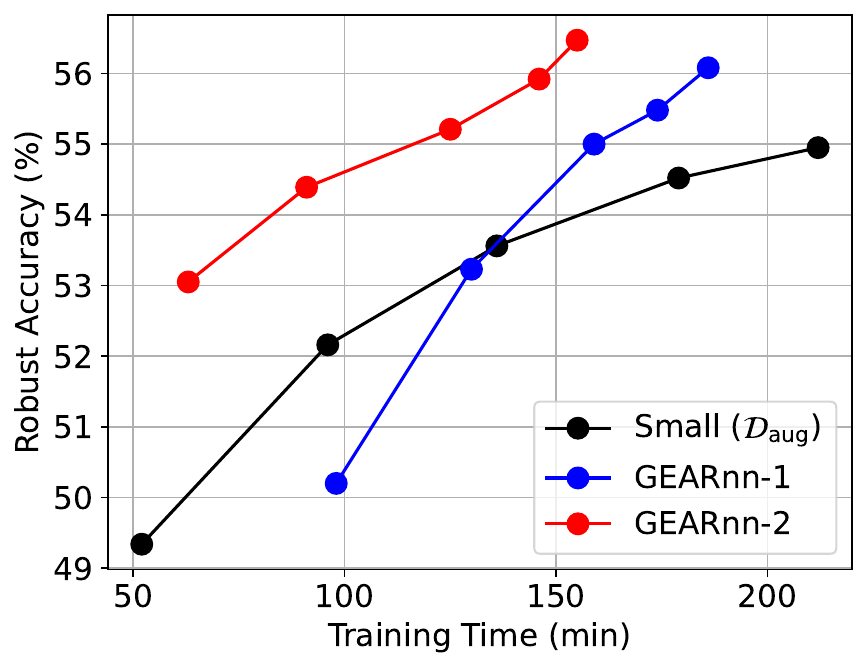}
  \subcaption{\label{c100 robustness}}
\end{subfigure}
\begin{subfigure}{.32\textwidth}
  \centering
  \includegraphics[width=\linewidth]{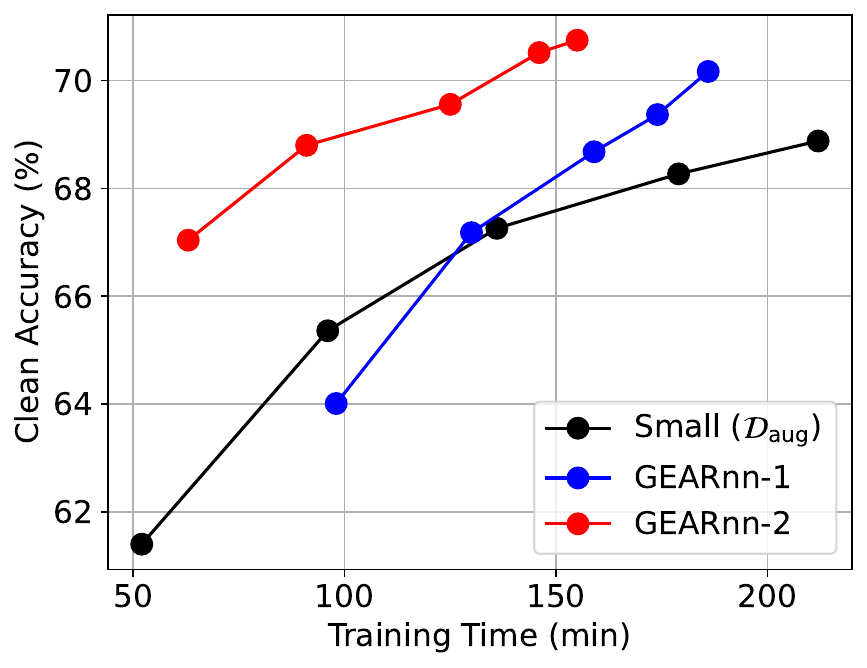}
  \subcaption{\label{c100 traintime}}
\end{subfigure}
\caption{\label{fig: robust epochs vgg mob}Plots (a)-(c) are implemented for VGG-19/CIFAR-10, (d)-(f) are for MobileNet-V1/CIFAR-10, (g)-(i) are for VGG-19/CIFAR-100, and (j)-(l) are for MobileNet-V1/CIFAR-100 on Quadro. First two plots of each row indicates the robust accuracy as a function of epochs and training time respectively. The last plot in each row shows the clean accuracy as a function of training time. GEARnn-2 clearly achieves the best clean and robust accuracy at the same training cost.
\vspace{-10pt}}
\end{figure*}

In~\cref{sec: results} and~\cref{fig: robust epochs} we observed that GEARnn-2 can achieve high robustness even when the robust training epochs are low. This is due to better initialization provided by OSG. We show the same results ablated for both VGG-19 and MobileNet-V1 for CIFAR-10 and CIFAR-100 in~\cref{fig: robust epochs vgg mob}.

\clearpage
\end{document}